\documentclass{article} % For LaTeX2e
\usepackage{preprint,times}

\usepackage{amsmath,amsfonts,bm}

\def\eqref#1{equation~\ref{#1}}
\def\1{\bm{1}}

\DeclareMathAlphabet{\mathsfit}{\encodingdefault}{\sfdefault}{m}{sl}
\SetMathAlphabet{\mathsfit}{bold}{\encodingdefault}{\sfdefault}{bx}{n}

\usepackage{hyperref}
\usepackage{url}
\usepackage{booktabs}

\usepackage[table]{xcolor}
\usepackage{booktabs} % \toprule, \midrule, \bottomrule, \cmidrule

\usepackage{graphicx}  % \resizebox
\usepackage{tabularx}
\usepackage[most]{tcolorbox}
\usepackage{enumitem}
\usepackage{multirow}
\usepackage{hyperref}

\definecolor{gainblue}{RGB}{0,83,159}
\definecolor{lossorange}{RGB}{180,70,0}

\definecolor{casegray}{RGB}{125,125,125}
\definecolor{casegraybg}{RGB}{238,238,238}

\newtcolorbox{casebox}[1]{
    colback=casegraybg,
    colframe=casegray,
    colbacktitle=casegraybg,
    coltitle=black,
    boxrule=0.6pt,
    arc=2pt,
    left=6pt,
    right=6pt,
    top=5pt,
    bottom=5pt,
    title=\textbf{#1},
    fonttitle=\bfseries
}

\newcommand{\gain}[1]{{\color{gainblue}(+#1)}}
\newcommand{\loss}[1]{{\color{lossorange}(-#1)}}

\usepackage[most]{tcolorbox}
\tcbuselibrary{listings,breakable}

\definecolor{casegray}{RGB}{125,125,125}
\definecolor{casegraybg}{RGB}{238,238,238}

\newtcblisting{promptbox}[1]{
    listing only,
    breakable,
    colback=casegraybg,
    colframe=casegray,
    colbacktitle=casegraybg,
    coltitle=black,
    boxrule=0.6pt,
    arc=2pt,
    left=6pt,
    right=6pt,
    top=5pt,
    bottom=5pt,
    title=\textbf{#1},
    fonttitle=\bfseries,
    listing options={
        basicstyle=\normalfont\normalsize,
        breaklines=true,
        breakindent=0pt,
        columns=fullflexible,
        keepspaces=true,
        showstringspaces=false,
        xleftmargin=0pt,
        xrightmargin=0pt
    }
}

\title{From Structure to Validity: Mechanism-Guided Generation and Refinement of LLM Rubrics}

\author{%
  \begin{minipage}{0.97\textwidth}
    \centering
    \normalfont\normalsize
    Feifei~Qian\textsuperscript{1,2},
    Xuan~Chen\textsuperscript{2},
    Lixin~Cui\textsuperscript{3},
    Chaoyang~Mei\textsuperscript{2},\\[2pt]
    Fanlin~Meng\textsuperscript{2},
    Shuyao~Xiao\textsuperscript{1},
    Ming~Cui\textsuperscript{2},
    Junxi~Yin\textsuperscript{2},
    Lu~Bai\textsuperscript{1}\endgraf
    \vspace{5pt}
    {\small
      \textsuperscript{1}\,School of Artificial Intelligence, Beijing Normal University\\
      \textsuperscript{2}\,KE Holdings Inc.\\
      \textsuperscript{3}\,School of Information, Central University of Finance and Economics\endgraf
    }
    \vspace{4pt}
    {\footnotesize
      \mbox{\texttt{feifei\_qian@mail.bnu.edu.cn}}\quad
      \mbox{\texttt{chenxuan046@ke.com}}\quad
      \mbox{\texttt{bailu@bnu.edu.cn}}\endgraf
    }
  \end{minipage}%
}

\begin{document}

\maketitle
\lhead{\small Preprint}

\begin{abstract}

Rubrics have been widely used to evaluate and optimize language models on open-ended tasks. However, existing work primarily focuses on generating better rubrics, with limited understanding of what makes rubric design effective. In this work, we conduct a systematic mechanistic analysis of rubric design across criterion organization, criterion operationalization, and score aggregation. Our analysis shows that structural organization enables more comprehensive criterion construction, semantic specification alone does not ensure effective criterion-level judgments, and sophisticated score aggregation provides limited additional gains. Guided by these findings, we introduce Adaptive Rubric Tree Search with Hybrid refinement (ARTS-Hybrid), which combines adaptive tree construction with semantic-first, response-calibrated refinement. Experiments across benchmarks show that ARTS-Hybrid improves both end-to-end evaluation performance and intrinsic rubric quality. Further experiments suggest that, beyond evaluation, rubrics generated by ARTS-Hybrid provide effective signals for model optimization.

\end{abstract}

\section{Introduction}

Large language models (LLMs) have made remarkable progress in recent years and are increasingly applied to open-ended tasks~\citep{Shi2025DeepRA}. Unlike tasks with verifiable outcomes, such as mathematical reasoning and code generation, open-ended tasks often admit multiple valid responses and lack objective ground-truth signals. Consequently, reliable evaluation has become a critical bottleneck in the development of LLMs. LLM-as-a-Judge~\citep{gu2026survey,Zheng2023judge} has emerged as a scalable alternative to costly human evaluation, with many existing methods relying on a single holistic score to assess response quality. However, this scalar evaluation collapses multiple dimensions of response quality, such as correctness, completeness, and relevance, into a coarse-grained judgment, making it difficult to provide fine-grained feedback for model improvement. 

Inspired by the long-standing use of rubrics in educational assessment~\citep{brookhart2018appropriate}, rubric-based evaluation offers a structured alternative by decomposing overall response quality into explicit task-specific criteria. In the context of LLMs, rubrics are increasingly becoming an important tool for model evaluation~\citep{hashemi-etal-2024-llm} and optimization~\citep{gunjal2026rubrics}. Early rubric construction methods primarily rely on language models to generate evaluation criteria directly from task instructions~\citep{viswanathan2026checklists,lin2025wildbench}, often resulting in incomplete coverage and inconsistent levels of granularity. Subsequent contrastive methods derive more discriminative evaluation criteria by identifying differences between preferred and rejected responses~\citep{liu2026openrubrics,tan2025judgebench,DBLP:journals/corr/abs-2510-17314}. More recent work has introduced structural decomposition~\citep{gao2026qworld,shen2026rethinking} and iterative refinement~\citep{zhang2026chasing,li2026rubrichub} to further improve the coverage and reliability of the rubric. In parallel, online approaches dynamically update rubrics based on policy model rollouts~\citep{sheng2026reinforcing,shao2025dr}, allowing evaluation criteria to evolve with the capabilities of the policy model. However, whether rubrics are constructed offline or dynamically evolved with the policy model, existing methods typically rely on aggregate outcomes, such as preference accuracy or downstream rewards, to guide their optimization. Although these metrics quantify end-to-end performance on a given dataset, the mechanisms underlying rubric effectiveness remain largely unexplained. 
The effectiveness of a rubric is influenced by multiple interdependent factors, such as how the criteria are constructed and organized. Consequently, the observed gains cannot be reliably attributed to specific design choices, providing limited systematic guidance on how rubrics should be structured or refined. This highlights the need for a systematic mechanistic analysis to disentangle the effects of individual design choices.

In this paper, we move beyond end-to-end comparisons and dissect rubric effectiveness along three stages, including how evaluation criteria are structurally organized, how individual criteria are operationalized, and how criterion-level scores are aggregated. Through systematic mechanistic analyses, we uncover three main findings. First, the structural decomposition improves criterion coverage and supports more targeted diagnosis. Second, the semantically sound evaluation criteria do not necessarily yield discriminative and stable judgments, motivating the use of response-based calibration evidence. Third, with consistent criterion content and criterion-level judgments across settings, the alternative score aggregation strategies provide only limited additional gains.

Guided by these findings, we introduce \textbf{A}daptive \textbf{R}ubric \textbf{T}ree \textbf{S}earch with \textbf{Hybrid} refinement (\textbf{ARTS-Hybrid}), a mechanism-guided framework that formulates rubric construction as a constrained search
over hierarchical criterion structures. Specifically, ARTS-Hybrid constructs rubrics through structured decomposition, where each expansion follows a semantic-first, response-calibrated selection process to identify the child criteria. We evaluate ARTS-Hybrid in terms of end-to-end evaluation performance and intrinsic rubric quality. In addition, we examine whether its benefits can be transferred to a standalone rubric generator and whether the generated rubrics provide effective reward signals for policy optimization. Our contributions are summarized as follows:
\begin{itemize}
    \item \textbf{A systematic mechanistic analysis of rubric effectiveness.}
    We conduct a systematic analysis of rubric design by disentangling three interdependent factors, including criterion organization, criterion operationalization, and score aggregation. 

    \item \textbf{A mechanism-guided generation and refinement framework.} We translate these findings into ARTS-Hybrid, which integrates structured decomposition with semantic-first, response-calibrated refinement. Experimental results show that ARTS-Hybrid achieves competitive performance across benchmarks while producing high-quality rubrics.

    \item \textbf{Validation beyond rubric-based evaluation.} We further show that the benefits of ARTS-Hybrid can be transferred to a standalone rubric generator and the generated rubrics provide effective reward signals for policy optimization.
    
\end{itemize}

\section{Related Work}

\textbf{Automatic rubric generation and refinement.} Existing approaches to automatic rubric construction can be broadly categorized into three paradigms according to how rubrics are updated: one-shot generation, offline iterative refinement, and online co-evolution. One-shot methods generate evaluation criteria directly from each task instruction~\citep{viswanathan2026checklists,lin2025wildbench}, with some methods additionally conditioning on candidate responses~\citep{gunjal2026rubrics}. Contrastive variants~\citep{liu2026openrubrics,tan2025judgebench,DBLP:journals/corr/abs-2510-17314} further compare preferred and rejected responses to identify the dimensions that distinguish stronger outputs from weaker ones, yielding more discriminative criteria. Despite leveraging richer response evidence, one-shot methods still construct rubrics in a single pass, with limited opportunities to correct missing or redundant criteria. This has motivated the development of offline iterative refinement, which progressively validates and revises the generated rubrics. Among them, RRD~\citep{shen2026rethinking} and RubricHub~\citep{li2026rubrichub} adopt coarse-to-fine refinement strategies, decomposing broad criteria into finer-grained, actionable evaluation items to improve coverage and applicability. Q-World~\citep{gao2026qworld} combines hierarchical and lateral expansion to explore a broader evaluation space across different levels of abstraction, task scenarios, and evaluation perspectives. These methods demonstrate the empirical benefits of hierarchical decomposition in rubric construction. Concurrent work~\citep{yang2026rubrics} has also explored tree-structured rubric construction, further extending hierarchical decomposition into an explicit structural representation. Beyond offline refinement, more recent online approaches update rubrics throughout policy optimization using newly generated rollouts~\citep{sheng2026reinforcing,shao2025dr,DBLP:journals/corr/abs-2608-29856}, allowing the evaluation criteria to evolve with the policy.

\textbf{Rubrics as rewards for LLM optimization.}
Rubrics can serve not only as evaluation instruments but also as reward functions for open-ended tasks. Rubrics as Rewards (RaR)~\citep{gunjal2026rubrics} constructs instance-specific criteria for each task and uses an LLM to judge whether a policy response satisfies each criterion. The resulting criterion-level scores are then aggregated into a scalar reward to guide policy optimization, extending reinforcement learning to open-ended tasks without programmatically verifiable outcomes. Subsequent work further improves the adaptability of rubric-based rewards. InfiMed-ORBIT~\citep{wang2025infimed} dynamically generates case-conditioned criteria for medical dialogues and employs them in incremental reinforcement learning, whereas DR-Tulu~\citep{shao2025dr} continually updates rubrics using rollouts from the current policy, allowing the reward specification to co-evolve with model capabilities and emerging failure modes. Meanwhile, OptimSyn~\citep{fan2026optimsyn} optimizes rubrics from the perspective of training utility, using influence estimates of synthetic examples on the target model as feedback to align rubric optimization more directly with downstream performance.

Overall, existing studies have substantially improved rubric generation and refinement. However, they primarily focus on how to construct or update better rubrics, while the mechanisms underlying their effectiveness remain underexplored. In particular, it remains unclear how structural decomposition, criterion-level validity, and score aggregation each contribute to overall rubric effectiveness.

\section{What Makes an Effective Rubric? A Mechanistic Analysis}
To address this gap, we conduct a systematic mechanistic analysis of rubric effectiveness centered on three research questions. \textbf{RQ1:} How does structural organization affect rubric quality? \textbf{RQ2:} What makes an individual criterion effective in practice? \textbf{RQ3:} Does score aggregation affect rubric evaluation? We construct a controlled analysis set of 1,500 instances through stratified sampling from HealthBench~\citep{arora2025healthbench} and RubricBench~\citep{zhou2026rubricbench}. Both datasets provide expert-authored atomic evaluation criteria, which we treat as gold rubric leaves. For each query, we manually construct and validate a \textbf{gold rubric tree} by recursively grouping semantically related leaves into higher-level evaluation dimensions. Additional details on the datasets and sampling procedure are provided in Appendix~\ref{app:data}. 

\subsection{RQ1: How Does Structural Organization Affect Rubric Quality?}
To answer RQ1, we examine two potential roles of structural organization in rubric-based evaluation. We investigate whether it helps the generator discover evaluation criteria more systematically and whether the resulting hierarchy provides useful context for judging individual criteria.

\textbf{Structural organization for criterion generator.} We first examine whether structural organization enables more systematic exploration of the evaluation space, rather than merely increasing the number of generated criteria. We compare tree-structured and flat rubric generation using the same generator. The hierarchical approach recursively constructs a rubric tree through structured decomposition, whereas the flat approach directly generates a set of criteria without an explicit hierarchy. To control for the additional generator calls introduced by recursive decomposition, we further include \textsc{Flat-Multi}, which independently runs flat generation the same number of times as the corresponding tree construction for each instance and reports the average performance. We use GLM-4.7-Flash as the generator and align the generated criteria with expert-annotated gold rubrics to measure coverage and novelty, with metric definitions provided in Appendix~\ref{app:metrics}. We use DeepSeek-V4-Flash-GA and Llama3.1-405B-Instruct as two independent judges.

\begin{table}[ht]
\caption{Comparison between tree-structured and flat rubric generation.}
\label{tab:tree_flat_comparison}
\centering
\begin{tabular}{ccccccc}
\toprule
\multirow{2}{*}{Judge Model} &
\multicolumn{3}{c}{Coverage (\%) $\uparrow$} &
\multicolumn{3}{c}{Novelty (\%) $\uparrow$} \\
\cmidrule(lr){2-4}
\cmidrule(lr){5-7}
&
\textsc{Flat} & \textsc{Flat-Multi} & \textsc{Tree} &
\textsc{Flat} & \textsc{Flat-Multi} & \textsc{Tree} \\
\midrule

DeepSeek-V4-Flash-GA
& 52.17 & 56.78 & \textbf{57.39}
& 9.69 & 9.16 & \textbf{11.46} \\

Llama3.1-405B-Instruct
& 42.11 & 41.13 & \textbf{43.90}
& 22.37 & 25.09 & \textbf{27.49} \\

\bottomrule
\end{tabular}
\end{table}

As shown in Table~\ref{tab:tree_flat_comparison}, tree-structured rubric generation outperforms both \textsc{Flat} and \textsc{Flat-Multi} in criterion coverage and valid novelty across judge models. Notably, the gains persist when the number of generator calls is matched, indicating that they cannot be explained by additional generation alone. These results suggest that hierarchical decomposition provides a more systematic scaffold for exploring the evaluation space and recovering fine-grained criteria that flat generation may overlook.

\paragraph{Hierarchical context for criterion judgment.} To examine whether hierarchical structure provides useful semantic context for criterion-level judgments, we construct controlled counterfactual responses based on the gold rubric trees. We first verify whether the reference response satisfies all gold criteria and revise it when necessary to obtain a canonical correct response. Starting from this response, we apply minimal local edits to construct three types of counterfactuals with increasing structural scope: (1) single-leaf, violating one gold leaf criterion; (2) within-dimension multi-leaf, jointly violating multiple leaves under the same semantic dimension; and (3) cross-dimension multi-leaf, introducing violations across multiple semantic dimensions. Detailed editing operations are provided in Appendix~\ref{app:counterfactual}. We formulate this evaluation as a controlled criterion-level violation detection task. Given a query, a counterfactual response, and its gold rubric, the judge model determines whether each rubric leaf is satisfied. The predefined intervention targets serve as ground-truth labels. We evaluate the performance using exact-match accuracy, i.e., a prediction is correct only if the judge model identifies all and only the violated leaves. We compare three ways of using rubric structure during judgment.
\textsc{Flat} evaluates each leaf independently without hierarchical context. \textsc{Tree} uses the rubric hierarchy for top-down routing, selecting potentially violated branches before making leaf-level judgments. \textsc{Path} also draws on the rubric tree but removes intermediate routing decisions, instead providing the full root-to-leaf semantic path when evaluating every criterion.

\begin{table}[ht]
\caption{Criterion-level violation detection under different uses of
rubric hierarchy. All results are reported as percentages, and colored
values indicate absolute changes from Flat in percentage points.}
\label{tab:counterfactual_tree_comparison}
\centering
\setlength{\tabcolsep}{4pt}
\begin{tabular}{ccccc}
\toprule
\multirow{2}{*}{Judge Model} & \multirow{2}{*}{Counterfactual Type} &
\multicolumn{3}{c}{Exact-Match Accuracy (\%)$\uparrow$} \\
\cmidrule(lr){3-5}
& & \textsc{Flat} & \textsc{Tree} & \textsc{Path} \\
\midrule
DeepSeek-V4-Flash-GA
& Single-leaf
& 26.84
& 26.78 \loss{0.06}
& \textbf{30.77} \gain{3.93} \\

DeepSeek-V4-Flash-GA
& Within-dimension
& 11.67
& 10.73 \loss{0.94}
& \textbf{12.21} \gain{0.54} \\

DeepSeek-V4-Flash-GA
& Cross-dimension
& 23.92
& 16.44 \loss{7.48}
& \textbf{28.18} \gain{4.26} \\

Llama3.1-405B-Instruct
& Single-leaf
& 33.91
& 32.35 \loss{1.56}
& \textbf{37.25} \gain{3.34} \\

Llama3.1-405B-Instruct
& Within-dimension
& 14.17
& 13.73 \loss{0.44}
& \textbf{16.46} \gain{2.29} \\

Llama3.1-405B-Instruct
& Cross-dimension
& 40.60
& 38.45 \loss{2.15}
& \textbf{41.94} \gain{1.34} \\
\bottomrule
\end{tabular}
\end{table}

As shown in Table~\ref{tab:counterfactual_tree_comparison}, the top-down \textsc{Tree} routing does not outperform \textsc{Flat} under either judge model. This does not imply that hierarchical structure is inherently ineffective. Instead, the parent-level routing introduces a decision bottleneck, i.e., if the judge fails to identify a violation at an internal node, the corresponding descendant leaves are excluded from further inspection, allowing a single routing error to obscure all violations within that branch. By contrast, \textsc{Path} consistently outperforms \textsc{Flat} across all six settings, suggesting that higher-level evaluation dimensions provide useful semantic context for interpreting the scope of individual criteria. Overall, hierarchical structure is more effective as a semantic scaffold for criterion judgment than as a direct basis for top-down pruning without explicit recall safeguards. 

To verify that the gains from \textsc{Path} arise from hierarchical semantics rather than merely increased context length, we introduce a \textsc{Shuffled-Path} control. Specifically, we replace the intermediate dimensions along each root-to-leaf path with semantically mismatched dimensions at the same depth, while keeping all other inputs unchanged. Detailed results are reported in Appendix~\ref{app:shuffled_path}. Having established the role of structural organization, we next shift from how criteria are organized to the effectiveness of the individual criteria themselves.

\subsection{RQ2: What Makes an Individual Criterion Effective in Practice?}
We characterize an effective criterion by two operational properties: \textbf{discriminative power} and \textbf{judge stability}. The former captures whether a criterion can distinguish responses of different quality, while the latter reflects whether its judgments remain consistent across judge models. We investigate how these properties can be realized in practice through two complementary mechanisms, \textbf{semantic specification} and \textbf{response-based calibration}. We conduct this analysis on 750 preference instances from RubricBench~\citep{zhou2026rubricbench}. For each instance, we assess discriminative power and judge stability using the two responses provided by the dataset and two independent judge models. We measure discriminative power by the variance of criterion scores across two responses under a fixed judge, and judge stability by the variance in scores across judges for the same response and criterion. In addition, we construct a calibration panel of six responses spanning strong, medium, and weak quality levels for each query. These responses serve only as signals for candidate criterion selection and are not used in the final evaluation. Detailed construction and scoring procedures are provided in Appendix~\ref{app:calibration}.

\begin{table}[t]
\caption{Effects of semantic specification and response-based calibration on criterion-level properties and pairwise evaluation accuracy, evaluated
using Llama-3.3-70B-Instruct and Qwen2.5-72B-Instruct~\citep{qwen25technicalreport} as judge models. \textit{Invalid} denotes the percentage of selected candidate
revisions that fail the semantic-validity constraint and is not
applicable to \textsc{Direct}.}
\label{tab:rq2_semantic_calibration}
\centering
\begin{tabular}{lcccccc}
\toprule
\multirow{2}{*}{Method}
& \multicolumn{3}{c}{Criterion-Level Properties}
& \multicolumn{3}{c}{Pairwise Accuracy (\%)} \\
\cmidrule(lr){2-4}
\cmidrule(lr){5-7}
&
Disc. $\uparrow$
& Judge Var. $\downarrow$
& Invalid (\%) $\downarrow$
& Llama
& Qwen
& Avg. $\uparrow$ \\
\midrule
Direct
& 0.5105
& 0.2749
& --
& 48.27
& 47.87
& 48.07 \\

Semantic
& 0.5207
& 0.2726
& \textbf{0.00}
& \textbf{49.87}
& 47.00
& 48.44 \\
\midrule
Calibration
& \textbf{0.5450}
& \textbf{0.2510}
& 17.84
& 48.27
& 49.73
& 49.00 \\

\rowcolor{gray!10}
Semantic + Calibration
& 0.5295
& 0.2626
& \textbf{0.00}
& 49.33
& \textbf{50.53}
& \textbf{49.93} \\
\bottomrule
\end{tabular}
\end{table}

\textbf{Can semantic specification induce operational effectiveness?}
We first examine whether specifying the desired properties during criterion generation is sufficient to realize them in practice. Specifically, we compare two settings, \textsc{Direct} and \textsc{Semantic}. \textsc{Direct} generates criteria without additional constraints, whereas \textsc{Semantic} explicitly encourages criteria that are semantically valid, discriminative, and judge-stable. We assess the resulting criteria in terms of discriminative power, cross-judge variance, and pairwise evaluation accuracy. As shown in Table~\ref{tab:rq2_semantic_calibration}, semantic specification shifts both operational properties \textbf{in the desired direction, but only marginally}. This reveals a gap between specifying desirable criterion properties and realizing them in actual evaluation behavior.

\begin{table}[ht]
\caption{
Pairwise evaluation accuracy under different score aggregation schemes.
}
\label{tab:rq3_weighting}
\centering
\setlength{\tabcolsep}{7pt}
\begin{tabular}{lcc}
\toprule
Aggregation Scheme
& DeepSeek-V4-Flash-GA (\%) $\uparrow$
& Llama3.1-405B-Instruct  (\%) $\uparrow$ \\
\midrule

\textsc{Flat}
& \textbf{73.20}
& 73.00 \\

\textsc{Tree}
& 72.60 \loss{-0.60}
& 71.33 \loss{-1.67} \\

\textsc{Flat + WU}
& 72.20 \loss{-1.00}
& \textbf{73.27} \gain{0.27} \\

\textsc{Tree + WU}
& 72.53 \loss{-0.67}
& 71.33 \loss{-1.67} \\

\bottomrule
\end{tabular}
\end{table}

\textbf{Grounding criterion effectiveness through response-based calibration.} The limited gains from semantic specification motivate us to introduce empirical evidence from actual model responses. Rather than assessing criterion quality solely from its semantic description, response-based calibration evaluates how candidate criteria behave on a panel of calibration responses, providing direct evidence of their discriminative power and judge stability. We compare two settings, \textsc{Calibration} and \textsc{Semantic+Calibration}, both operating on the same fixed pool of local rubric revisions. The former selects candidates based on their behavioral characteristics on calibration responses, whereas the latter applies the same utility only after semantic filtering. Detailed definitions are provided in Appendix~\ref{app:rq2_calibration}. As shown in Table~\ref{tab:rq2_semantic_calibration}, \textsc{Calibration} achieves the highest discriminative power and the lowest cross-judge variance. However, 17.84\% of the selected criteria are semantically invalid, indicating that operational effectiveness alone does not ensure alignment with the underlying task requirements. In contrast, applying semantic constraints before calibration eliminates invalid selections while retaining most of the gains in discrimination and stability. Consequently, \textsc{Semantic+Calibration} achieves the highest average pairwise evaluation accuracy. These results show that semantic constraints are necessary to prevent task-misaligned criteria, while response-based calibration is needed to identify criteria that are discriminative and stable in practice.

\subsection{RQ3: Does Score Aggregation Affect Rubric Evaluation?}
RQ1 and RQ2 examine the roles of criterion organization and criterion-level effectiveness in rubric-based evaluation respectively. We now turn to the final stage, where criterion-level judgments are aggregated into an overall score. By fixing the criterion content and varying only the aggregation strategy, we assess whether weighting and hierarchical aggregation contribute additional gains. For the \textsc{Flat} setting, the rubric generator independently assigns each leaf $l$ an importance score $a_l\in\{1,...,5\}$, which is normalized across all leaves as $w_l = a_l / \sum_{k \in \mathcal{L}} a_k$, where \(\mathcal{L}\) denotes the set of leaf criteria. 
\textsc{Flat+WU} applies the global weighting scheme introduced in RRD~\citep{shen2026rethinking} to leaf-level scores before aggregation, while \textsc{Tree+WU} applies the same weighting locally within the hierarchy. All methods share identical rubric leaves and criterion-level judgments, differing only in how leaf scores are weighted and aggregated.

As shown in Table~\ref{tab:rq3_weighting}, neither hierarchical aggregation nor WU weighting yields consistent improvements. With DeepSeek-V4-Flash-GA as the judge, \textsc{Flat} achieves the highest accuracy, while \textsc{Flat+WU} provides only a marginal gain of 0.27 percentage points under Llama3.1-405B-Instruct. Overall, once criterion-level evaluation is established, the final aggregation strategy provides limited additional gains, suggesting that rubric effectiveness depends more on how criteria are designed.

\section{ARTS-hybrid: Mechanism-Guided Rubric Generation and Refinement}
\begin{figure}[ht]
    \centering
    \includegraphics[width=0.85\linewidth]{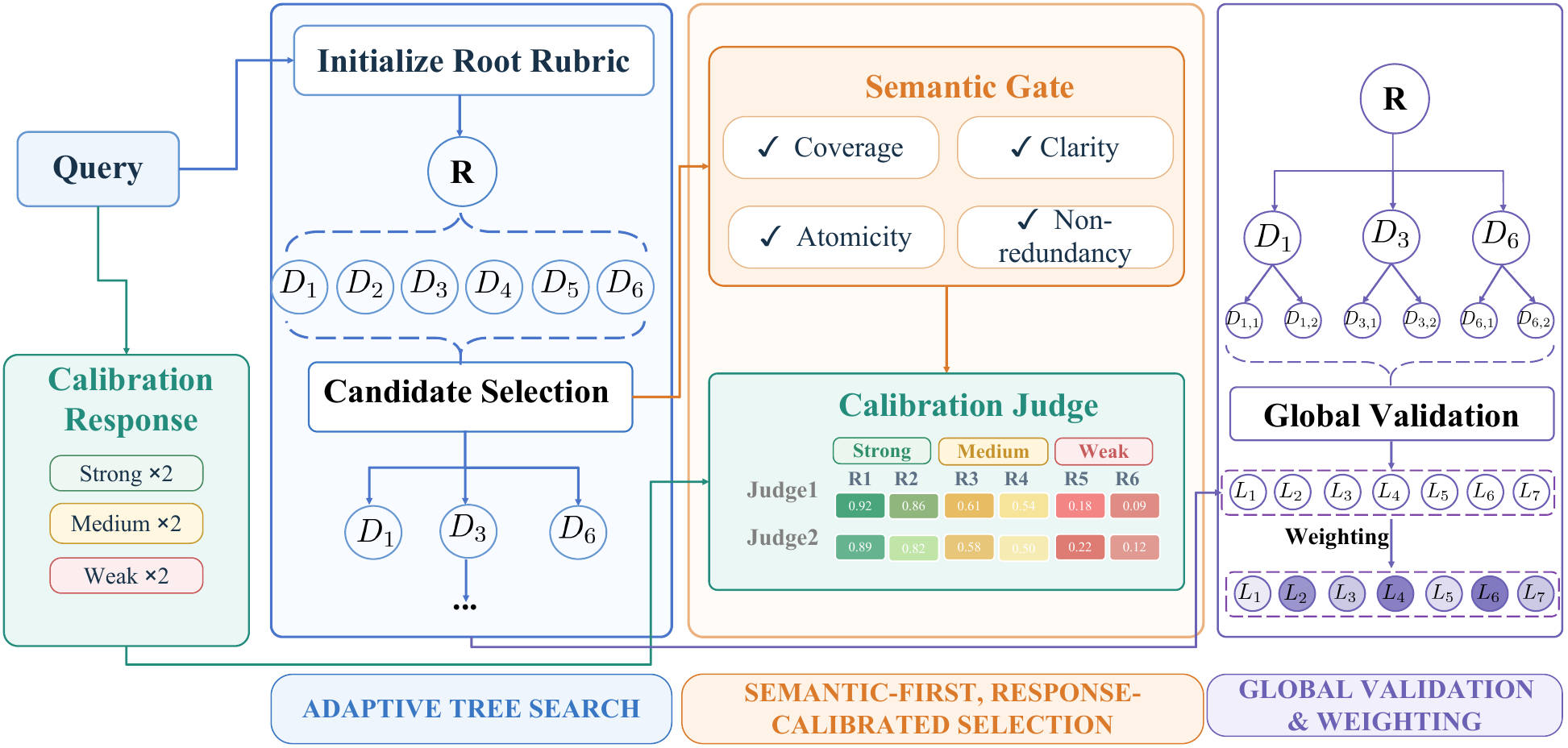}
    \caption{Overview of the ARTS-Hybrid framework.}
    \label{fig:arts_framework}
\end{figure}
Building on the above mechanistic findings, we introduce \textbf{ARTS-Hybrid}, a rubric generation and refinement framework that translates these insights into a principled construction process. Given a query, ARTS-Hybrid progressively constructs hierarchical evaluation criteria through adaptive tree search. At each node expansion, semantic constraints first identify feasible candidate decompositions, while calibration responses provide empirical evidence to guide the final selection. Once the search terminates, ARTS-Hybrid performs global validation and refinement of the final leaf criteria, followed by normalized weight assignment to produce the final rubric. We next describe the three key components of ARTS-Hybrid in detail, including adaptive tree construction, constrained candidate selection, and global validation with weight assignment.

\subsection{Adaptive Rubric Tree Construction}
Motivated by the findings from RQ1, we adopt a hierarchical tree structure for adaptive rubric construction. Given a query, ARTS-Hybrid first initializes a root rubric representing overall response quality. The root serves only as a virtual structural node for organizing subsequent decomposition and does not directly contribute to the final score. Rather than expanding the tree in a predetermined order, ARTS-Hybrid maintains unprocessed nodes in a priority queue and performs best-first search according to their importance and depth. For each node \(v\) at depth \(d_v\), we independently estimate an importance score \(I(v)\in\{1,\ldots,5\}\) according to its relevance to the current task. The expansion priority is defined as
\begin{equation}
P(v) = I(v)\gamma^{d_v},
\label{eq:node_priority}
\end{equation}
where \(\gamma\in(0,1]\) denotes the depth discount factor. At each iteration, ARTS-Hybrid selects the highest-priority unexpanded node from the queue and determines whether it is sufficiently atomic and independently assessable to serve as a leaf criterion. Nodes that satisfy these conditions are finalized as leaves. Otherwise, the rubric generator produces multiple finer-grained child criteria, with the selected children added to the rubric tree and priority queue for further expansion. This process continues until no expandable nodes remain or a predefined stopping condition is reached.

\subsection{Semantic-First Response-Calibrated Selection}
Once an expandable node is selected, ARTS-Hybrid selects the most suitable decomposition from the generated candidate set. Following the findings from RQ2, semantic constraints and response-based calibration play complementary roles. Accordingly, ARTS-Hybrid adopts a semantic-first, response-calibrated selection strategy. 

\paragraph{Semantic feasibility.} For each expandable node $v$, the rubric generator first produces a pool of candidate child criteria. These criteria are then combined into multiple candidate decompositions $\mathcal{C}_v$, where each $C\in\mathcal{C}_v$ represents a possible set of children for node $v$. ARTS-Hybrid applies a multi-stage semantic validation procedure to determine which decompositions are feasible with respect to the query $q$ and the current rubric tree $T$. The validation considers several aspects of rubric quality, including criterion clarity, task relevance, independent assessability, coverage of parent requirements, and semantic non-redundancy among child criteria. Only candidate decompositions that pass all semantic checks are retained in the feasible set
\begin{equation}
\mathcal{F}_v = \left\{ C\in\mathcal{C}_v \;\middle|\; \operatorname{SemValid}(C,q,T)=1 \right\},
\end{equation}
where \(\operatorname{SemValid}(C,q,T)\) indicates whether candidate decomposition \(C\) satisfies all semantic constraints with respect to the query \(q\) and the current rubric tree \(T\). Semantically invalid candidates are filtered out at this stage and do not proceed to subsequent response-based calibration.

\paragraph{Response-based calibration.} As demonstrated in RQ2, semantic information alone is insufficient to determine how effectively a criterion operates on actual model responses. We therefore use response-based calibration to provide empirical evidence for selecting among semantically valid decompositions. For each query, we construct a panel of six calibration responses spanning strong, medium, and weak quality levels and reuse it throughout the tree search. For every semantically feasible candidate, two independent calibration judges evaluate each criterion across six calibration responses, yielding a \(6\times2\) response-judge score matrix.
This matrix captures criterion behavior along two complementary dimensions. The variation across responses reflects the criterion's discriminative power, whereas variation across judges captures the consistency of its scoring behavior. We further compare response-score profiles across child criteria to identify behavioral redundancy. Based on these signals, we define the following calibration objective for each feasible candidate decomposition \(C\) as
\begin{equation}
J(C)
=
\alpha_D D(C)
+
\alpha_A A(C)
+
\alpha_S S(C)
-
\lambda_{\mathrm{red}}R(C),
\label{eq:calibration_objective}
\end{equation}
where \(D(C)\) measures response discrimination, \(A(C)\) and \(S(C)\) capture complementary aspects of cross-judge consistency, and \(R(C)\) penalizes redundant scoring behavior among child criteria. Detailed definitions of these terms are provided in Appendix~\ref{app:calibration_objective}. The final decomposition for node $v$ is then selected from the semantically feasible set by
\begin{equation}
C_v^{*}
=
\underset{C \in \mathcal{F}_v}{\arg\max}\; J(C).
\label{eq:candidate_selection}
\end{equation}
In this way, semantic validation determines which decompositions are admissible, while response-based calibration determines which admissible decomposition is most effective in practice.
\subsection{Global Validation and Weight Assignment}
While the above local search ensures semantic validity at each expansion step, locally valid decompositions do not necessarily yield a coherent and complete rubric. ARTS-Hybrid therefore performs global validation once the tree search is complete, ensuring full coverage of the evaluation requirements, leaf-level atomicity, and semantic non-redundancy.
When global validation identifies remaining issues, ARTS-Hybrid refines the leaf set by recovering missing evaluation requirements and restructuring existing criteria when necessary.
% The global validation is repeated after each refinement round until all constraints are satisfied or a predefined maximum number of iterations is reached. 
With the leaf set established, ARTS-Hybrid assigns a normalized weight to each criterion. Motivated by the findings from RQ3, we avoid propagating weights through the rubric hierarchy and instead estimate the importance of each leaf directly. Specifically, for each leaf $l$, the rubric generator assigns an importance score $a_l\in\{1,\ldots,5\}$ based on the query and the criterion itself. 

\section{Evaluation of ARTS-Hybrid}
We evaluate ARTS-Hybrid in terms of end-to-end evaluation performance and intrinsic rubric quality. In addition, we examine whether its benefits can be transferred to a standalone rubric generator and extended to policy optimization through rubric-based rewards.

\subsection{Experimental Setup}
\textbf{Benchmarks and baselines.} We evaluate on preference benchmarks across diverse domains and difficulty levels, including JudgeBench~\citep{tan2025judgebench}, RM-Bench Chat~\citep{Liurmbench}, the Chat and Chat-Hard subsets of RewardBench~\citep{lambert2025rewardbench}, the Precise-IF and Focus subsets of RewardBench 2~\citep{malik2026rewardbench2}, and RubricBench~\citep{zhou2026rubricbench}. Each instance contains a query, a pair of candidate responses, and a preference label. We report pairwise accuracy against human preferences as the primary metric. We compare ARTS-Hybrid with Direct Judge, Query-Only, Response-Aware, and Q-World~\citep{gao2026qworld}. These baselines represent direct evaluation, query-only rubric generation, response-aware rubric generation, and structured rubric refinement, respectively. We reproduce all baselines under the same evaluation protocol using the same generator and judge models. Detailed implementation settings are provided in Appendix~\ref{app:implementation}.

% To ensure a fair comparison, all rubric-based methods use the same rubric generator unless otherwise specified. The resulting rubrics are evaluated using the same judge models and prompting protocol across methods. For each response pair, the judge independently assesses the candidate responses against the generated criteria, and the criterion-level judgments are aggregated into a pairwise preference prediction.

% \textbf{Implementation details.}
% We use GLM-4.7-Flash as the common rubric generator and judge model across methods. The ARTS-Hybrid follows the same response construction and calibration protocol as in RQ2, including the calibration panel and judge models. Full construction and scoring details are provided in Appendix~\ref{app:calibration}.

\subsection{Main Results}

\textbf{End-to-end evaluation performance.}
Table~\ref{tab:main-evaluation-results} reports pairwise evaluation accuracy across six benchmark settings. ARTS-Hybrid achieves the highest accuracy on five of the six settings and remains competitive on PreciseIF. In particular, ARTS-Hybrid improves over both direct judgment and one-shot rubric generation on most benchmarks, while also outperforming Q-World on five settings. These results suggest that the mechanism-guided construction process provides a robust basis for rubric-based evaluation across diverse preference tasks.

\begin{table}[t]
\caption{
Pairwise accuracy (\%) across benchmarks. The best result is highlighted in bold.
}
\label{tab:main-evaluation-results}
\centering
\setlength{\tabcolsep}{5pt}
\begin{tabular}{@{}lccccc@{}}
\toprule
Benchmark
& Direct Judge
& Query-Only 
& Response-Aware 
& Q-World
& ARTS-Hybrid (ours)\\
\midrule

JudgeBench
& 63.14
& 58.57
& 61.43
& 63.71
& \textbf{64.29} \\

\midrule
RM-Bench Chat
& 61.24
& 60.47
& 59.69
& 62.79
& \textbf{64.34} \\

RB Chat
& 87.15
& 82.40
& 86.87
& 81.28
& \textbf{89.39} \\

Chat-Hard
& 58.11
& 64.47
& 59.87
& 65.79
& \textbf{68.20} \\

PreciseIF
& 52.92
& \textbf{56.25}
& 55.00
& 56.04
& 55.63 \\

Focus
& 79.80
& 80.27
& 84.44
& 83.03
& \textbf{85.59} \\

\bottomrule
\end{tabular}
\end{table}

\textbf{Intrinsic rubric quality.}
We further evaluate the intrinsic quality of the generated rubrics using DeepSeek-V4-Flash as an independent evaluator, with results from an additional evaluator provided in Appendix~\ref{app:intrinsic_quality}. As shown in Table~\ref{tab:intrinsic-rubric-quality}, ARTS-Hybrid achieves the highest scores across all four quality dimensions, including coverage, concreteness, non-redundancy, and actionability. This result complements the end-to-end evaluation results by showing that the gains are accompanied by improvements in the quality of the generated criteria themselves.
\begin{table}[ht]
\caption{
Comparison of intrinsic rubric quality.
Higher values indicate better performance.
}
\label{tab:intrinsic-rubric-quality}
\centering
\setlength{\tabcolsep}{3pt}
\begin{tabular}{lccccc}
\toprule
Method
& Coverage
& Concreteness
& Non-red.
& Actionability
& Overall \\
\midrule
Query-Only
& 1.926
& 2.720
& 2.349
& 2.871
& 2.049 \\

Response-Aware
& 1.966
& 2.677
& 2.329
& 2.734
& 2.029 \\

Q-World
& 2.082
& 2.434
& 1.041
& 2.222
& 1.997 \\

ARTS-Hybrid (ours)
& \textbf{2.163}
& \textbf{2.871}
& \textbf{2.851}
& \textbf{2.929}
& \textbf{2.206} \\
\bottomrule
\end{tabular}%
\end{table}

\textbf{Rubric generator distillation.}
The ARTS-Hybrid incurs additional computation primarily during offline rubric construction. We therefore examine whether the benefits of mechanism-guided construction can be distilled into a standalone rubric generator, avoiding repeated execution of the full ARTS-Hybrid pipeline. Specifically, we fine-tune the same Qwen3-4B rubric generator using three sources of supervision, including direct-prompt rubrics, expert-provided rubrics from the original dataset, and the rubrics generated by our framework, with the untrained Qwen3-4B generator as the baseline. As shown in Table~\ref{tab:sft-rubric-generator}, training on ARTS-Hybrid rubrics achieves the highest accuracy on JudgeBench. These results suggest that the benefits obtained through offline mechanism-guided construction can be transferred to a standalone rubric generator, providing a potential way to reduce reliance on the full ARTS-Hybrid pipeline for subsequent rubric generation.

\begin{table}[ht]
\vspace{-7pt}
\caption{
The performance of Qwen3-4B rubric generators under different SFT supervision sources.
}
\label{tab:sft-rubric-generator}
\centering
\setlength{\tabcolsep}{7pt}
\begin{tabular}{lc}
\toprule
SFT Supervision & JudgeBench Accuracy (\%)$\uparrow$ \\
\midrule
Qwen3-4B
& 69.14 \\

Direct-Prompt Rubrics
& 66.86 \\

OpenRubrics Expert Rubrics
& 66.57 \\

Qworld Rubrics & 64.86 \\

ARTS-Hybrid Rubrics
& \textbf{69.43} \\
\bottomrule
\end{tabular}
\vspace{-7pt}
\end{table}

\textbf{Rubric-as-Reward for policy optimization.}
Beyond evaluation, we examine whether ARTS-Hybrid can provide effective reward signals for policy optimization. We stratifiedly sample 3,000 training instances from OpenRubrics~\citep{liu2026openrubrics} and optimize Qwen3-4B using Dr.GRPO~\citep{liu2025understanding}. We compare reward signals derived from the same three rubric sources described above and additionally report the untrained Qwen3-4B model as the base model. Table~\ref{tab:rubric-reward-healthbench} reports the performance under both thinking and non-thinking settings. The ARTS-Hybrid achieves the best results in both settings, while directly generated and expert-provided rubrics provide limited gains over the base model. These results suggest that the advantages of ARTS-Hybrid are not limited to evaluation, but can also translate into more effective reward signals for policy optimization.

\begin{table}[ht]
\vspace{-7pt}
\caption{
The score performance on HealthBench under different rubric-based rewards. }
\label{tab:rubric-reward-healthbench}
\centering
\setlength{\tabcolsep}{5pt}
\begin{tabular}{lcc}
\toprule
Reward Source
& Thinking
& Non-Thinking \\
\midrule
Qwen3-4B 
& 38.14
& 34.86 \\

Direct-Prompt Rubric
& 38.08
& 34.76 \\

OpenRubrics Expert Rubric
& 38.33
& 34.85 \\

Qworld Rubric 
& 37.64
& 34.95 \\

ARTS-Hybrid Rubric (ours)
& \textbf{38.93}
& \textbf{35.18} \\
\bottomrule
\end{tabular}
\end{table}

\textbf{Additional experiments.}
We conduct two additional experiments in Appendix~\ref{app:additional_results}. (i) \textbf{Cross-judge robustness.} We evaluate the generated rubrics with different judge models to assess whether their effectiveness generalizes across evaluators. (ii) \textbf{Ablation study.} We conduct an ablation study to examine the contributions of semantic validation and response-based calibration.

\section{Conclusion}
In this work, we systematically investigate what makes rubric design effective from three perspectives, including criterion organization, criterion-level operationalization, and score aggregation. Guided by the resulting mechanistic findings, we introduce ARTS-Hybrid, which combines adaptive tree construction with semantic-first, response-calibrated refinement. Experiments show that ARTS-Hybrid produces higher-quality rubrics and achieves strong evaluation performance across different judge models. Beyond evaluation, our reinforcement learning experiments further show that well-designed rubrics can provide effective reward signals for model optimization.

\subsection*{Reproducibility Statement}
To support reproducibility, we release the manually constructed counterfactual responses and gold rubric trees used in the mechanistic analysis, together with the core code for both the mechanistic analysis and ARTS-Hybrid, in our code repository (\href{https://github.com/SophiaQian02/Tree_of_Rubrics_code}{https://github.com/SophiaQian02/Tree\_of\_Rubrics\_code}). The construction and refinement procedures of ARTS-Hybrid are described in the main text. Detailed implementation settings for the mechanistic analysis and ARTS-Hybrid are provided in the appendix. Appendix~\ref{app:data} describes the data sources and sampling procedure, while Appendix~\ref{app:counterfactual} details the construction of counterfactual responses. The response-based calibration setup, including the construction of calibration responses and the calibration judges, is provided in Appendix~\ref{app:calibration}. Appendix~\ref{app:params} summarizes the key hyperparameters and their search ranges. Finally, Appendix~\ref{app:prompts} provides the core prompts used in the mechanistic analysis and ARTS-Hybrid.



\bibliography{references}

@article{Shi2025DeepRA,
  title={Deep research: A systematic survey},
  author={Shi, Zhengliang and Chen, Yiqun and Li, Haitao and Sun, Weiwei and Ni, Shiyu and Lyu, Yougang and Fan, Run-Ze and Jin, Bowen and Weng, Yixuan and Zhu, Minjun and others},
  journal={arXiv preprint arXiv:2512.02038},
  year={2025}
}

@article{gu2026survey,
  title={A survey on llm-as-a-judge},
  author={Gu, Jiawei and Jiang, Xuhui and Shi, Zhichao and Tan, Hexiang and Zhai, Xuehao and Xu, Chengjin and Li, Wei and Shen, Yinghan and Ma, Shengjie and Liu, Honghao and others},
  journal={The Innovation},
  volume={7},
  number={6},
  year={2026},
  publisher={Elsevier}
}

@inproceedings{brookhart2018appropriate,
  title={Appropriate criteria: Key to effective rubrics},
  author={Brookhart, Susan M},
  booktitle={Frontiers in education},
  volume={3},
  pages={22},
  year={2018},
  organization={Frontiers Media SA}
}

@inproceedings{gunjal2026rubrics,
  title={Rubrics as rewards: Reinforcement learning beyond verifiable domains},
  author={Gunjal, Anisha and Wang, Anthony and Lau, Elaine and Nath, Vaskar and He, Yunzhong and Liu, Bing and Hendryx, Sean},
  booktitle={Proceedings of ICLR},
  pages={127924--127945},
  year={2026}
}

@inproceedings{hashemi-etal-2024-llm,
  title={LLM-rubric: A multidimensional, calibrated approach to automated evaluation of natural language texts},
  author={Hashemi, Helia and Eisner, Jason and Rosset, Corby and Van Durme, Benjamin and Kedzie, Chris},
  booktitle={Proceedings of ACL},
  pages={13806--13834},
  year={2024}
}

@inproceedings{viswanathan2026checklists,
  title={Checklists are better than reward models for aligning language models},
  author={Viswanathan, Vijay and Sun, Yanchao and Kong, Xiang and Cao, Meng and Neubig, Graham and Wu, Sherry},
  booktitle={Proceedings of NeurIPS},
  volume={38},
  pages={114728--114754},
  year={2026}
}

@inproceedings{lin2025wildbench,
  title={Wildbench: Benchmarking llms with challenging tasks from real users in the wild},
  author={Lin, Bill Yuchen and Deng, Yuntian and Chandu, Khyathi and Ravichander, Abhilasha and Pyatkin, Valentina and Dziri, Nouha and Le Bras, Ronan and Choi, Yejin},
  booktitle={Proceedings of ICLR},
  volume={2025},
  pages={47852--47870},
  year={2025}
}

@inproceedings{liu2026openrubrics,
  title={Openrubrics: Towards scalable synthetic rubric generation for reward modeling and llm alignment},
  author={Liu, Tianci and Xu, Ran and Yu, Tony and Hong, Ilgee and Yang, Carl and Zhao, Tuo and Wang, Haoyu},
  booktitle={Proceedings of ACL},
  pages={17417--17437},
  year={2026}
}

@inproceedings{tan2025judgebench,
  title={Judgebench: A benchmark for evaluating llm-based judges},
  author={Tan, Sijun and Zhuang, Siyuan and Montgomery, Kyle and Tang, William and Cuadron, Alejandro and Wang, Chenguang and Popa, Raluca and Stoica, Ion},
  booktitle={Proceedings of ICLR},
  volume={2025},
  pages={63277--63303},
  year={2025}
}

@article{DBLP:journals/corr/abs-2510-17314,
  author       = {Lipeng Xie and
                  Sen Huang and
                  Zhuo Zhang and
                  Anni Zou and
                  Yunpeng Zhai and
                  Dingchao Ren and
                  Kezun Zhang and
                  Haoyuan Hu and
                  Boyin Liu and
                  Haoran Chen and
                  Zhaoyang Liu and
                  Bolin Ding},
  title        = {Auto-Rubric: Learning to Extract Generalizable Criteria for Reward
                  Modeling},
  journal      = {CoRR},
  volume       = {abs/2510.17314},
  year         = {2025}
}

@article{gao2026qworld,
  title={Qworld: Question-specific evaluation criteria for llms},
  author={Gao, Shanghua and Su, Yuchang and Sui, Pengwei and Ginder, Curtis and Zitnik, Marinka},
  journal={arXiv preprint arXiv:2603.23522},
  year={2026}
}

@article{shen2026rethinking,
  title={Rethinking rubric generation for improving llm judge and reward modeling for open-ended tasks},
  author={Shen, William F and Qiu, Xinchi and Whitehouse, Chenxi and Alazraki, Lisa and Goel, Shashwat and Barbieri, Francesco and Willi, Timon and Mathur, Akhil and Leontiadis, Ilias},
  journal={arXiv preprint arXiv:2602.05125},
  year={2026}
}

@inproceedings{zhang2026chasing,
  title={Chasing the tail: Effective rubric-based reward modeling for large language model post-training},
  author={Zhang, Junkai and Wang, Zihao and Gui, Lin and Mysore Sathyendra, Swarnashree and Jeong, Jaehwan and Veitch, Victor and Wang, Wei and He, Yunzhong and Liu, Bing and Jin, Lifeng},
  booktitle={Proceedings of ICLR},
  volume={2026},
  pages={133430--133457},
  year={2026}
}

@inproceedings{li2026rubrichub,
  title={Rubrichub: A comprehensive and highly discriminative rubric dataset via automated coarse-to-fine generation},
  author={Li, Sunzhu and Zhao, Jiale and Ren, Huimin and Wei, Zhenlin and Zhou, Yang and Yang, Jingwen and Liu, Shunyu and Zhang, Kaike and Wei, Chen},
  booktitle={Proceedings of ACL},
  pages={31320--31344},
  year={2026}
}

@article{sheng2026reinforcing,
  title={Reinforcing chain-of-thought reasoning with self-evolving rubrics},
  author={Sheng, Leheng and Ma, Wenchang and Hong, Ruixin and Wang, Xiang and Zhang, An and Chua, Tat-Seng},
  journal={arXiv preprint arXiv:2602.10885},
  year={2026}
}

@article{shao2025dr,
  title={Dr tulu: Reinforcement learning with evolving rubrics for deep research},
  author={Shao, Rulin and Asai, Akari and Shen, Shannon Zejiang and Ivison, Hamish and Kishore, Varsha and Zhuo, Jingming and Zhao, Xinran and Park, Molly and Finlayson, Samuel G and Sontag, David and others},
  journal={arXiv preprint arXiv:2511.19399},
  year={2025}
}

@article{yang2026rubrics,
  title={Rubrics on Trial: Evolving Rubrics from a Single Query via Synthetic Pairwise Evidence},
  author={Yang, Haocheng and Pan, Licheng and Li, Xiaoxi and Chen, Zhichao and Zhang, Zhiheng and Lu, Yuan and Li, Haoxuan and Wang, Hao},
  journal={arXiv preprint arXiv:2607.15092},
  year={2026}
}

@article{wang2025infimed,
  title={Infimed-orbit: Aligning llms on open-ended complex tasks via rubric-based incremental training},
  author={Wang, Pengkai and Liu, Pengwei and Zuo, Qi and Sang, Zhijie and Xie, Congkai and Yang, Hongxia},
  journal={arXiv preprint arXiv:2510.15859},
  year={2025}
}

@article{fan2026optimsyn,
  title={Optimsyn: Influence-guided rubrics optimization for synthetic data generation},
  author={Fan, Zhiting and Chen, Ruizhe and Hu, Tianxiang and Peng, Ru and Huang, Zenan and Xu, Haokai and Chen, Yixin and Wu, Jian and Zhao, Junbo and Liu, Zuozhu},
  journal={arXiv preprint arXiv:2604.00536},
  year={2026}
}

@article{arora2025healthbench,
  title={Healthbench: Evaluating large language models towards improved human health},
  author={Arora, Rahul K and Wei, Jason and Hicks, Rebecca Soskin and Bowman, Preston and Qui{\~n}onero-Candela, Joaquin and Tsimpourlas, Foivos and Sharman, Michael and Shah, Meghan and Vallone, Andrea and Beutel, Alex and others},
  journal={arXiv preprint arXiv:2505.08775},
  year={2025}
}

@inproceedings{zhou2026rubricbench,
  title={RubricBench: Aligning Model-Generated Rubrics with Human Standards},
  author={Zhou, Junyi and Zhang, Qiyuan and Wang, Yufei and Lyu, Fuyuan and Ming, Yidong and Xu, Can and Sun, Qingfeng and Zheng, Kai and Kang, Peng and Liu, Xue and others},
  booktitle={Proceedings of ACL},
  pages={31179--31200},
  year={2026}
}

@inproceedings{Liurmbench,
author = {Liu, Dengcan and Yang, Fengkai and Wang, Xiaohan and Yan, Shurui and Chai, Jiajun and Li, Jiahao and Ban, Yikun and Mao, Zhendong and Lin, Wei and Yin, Guojun},
title = {CDRRM: Contrast-Driven Rubric Generation for Reliable and Interpretable Reward Modeling},
year = {2026},
booktitle = {Proceedings of ACM SIGKDD},
pages = {3152–3163},
numpages = {12}
}

@inproceedings{lambert2025rewardbench,
  title={Rewardbench: Evaluating reward models for language modeling},
  author={Lambert, Nathan and Pyatkin, Valentina and Morrison, Jacob and Miranda, LJ and Lin, Bill Yuchen and Chandu, Khyathi and Dziri, Nouha and Kumar, Sachin and Zick, Tom and Choi, Yejin and others},
  booktitle={Findings of the Association for Computational Linguistics: NAACL 2025},
  pages={1755--1797},
  year={2025}
}

@inproceedings{malik2026rewardbench2,
  title={Rewardbench 2: Advancing reward model evaluation},
  author={Malik, Saumya and Pyatkin, Valentina and Land, Sander and Morrison, Jacob and Smith, Noah and Hajishirzi, Hanna and Lambert, Nathan},
  booktitle={Proceedings of ICLR},
  volume={2026},
  pages={144839--144866},
  year={2026}
}

@inproceedings{
    liu2025understanding,
    title={Understanding R1-Zero-Like Training: A Critical Perspective},
    author={Zichen Liu and Changyu Chen and Wenjun Li and Penghui Qi and Tianyu Pang and Chao Du and Wee Sun Lee and Min Lin},
    booktitle={Proceedings of COLM},
    year={2025}
}

@article{yang2025qwen3,
  title={Qwen3 technical report},
  author={Yang, An and Li, Anfeng and Yang, Baosong and Zhang, Beichen and Hui, Binyuan and Zheng, Bo and Yu, Bowen and Gao, Chang and Huang, Chengen and Lv, Chenxu and others},
  journal={arXiv preprint arXiv:2505.09388},
  year={2025}
}

@article{grattafiori2024llama,
  title={The llama 3 herd of models},
  author={Grattafiori, Aaron and Dubey, Abhimanyu and Jauhri, Abhinav and Pandey, Abhinav and Kadian, Abhishek and Al-Dahle, Ahmad and Letman, Aiesha and Mathur, Akhil and Schelten, Alan and Vaughan, Alex and others},
  journal={arXiv preprint arXiv:2407.21783},
  year={2024}
}

@misc{jiang2023mistral7b,
      title={Mistral 7B}, 
      author={Albert Q. Jiang and Alexandre Sablayrolles and Arthur Mensch and Chris Bamford and Devendra Singh Chaplot and Diego de las Casas and Florian Bressand and Gianna Lengyel and Guillaume Lample and Lucile Saulnier and Lélio Renard Lavaud and Marie-Anne Lachaux and Pierre Stock and Teven Le Scao and Thibaut Lavril and Thomas Wang and Timothée Lacroix and William El Sayed},
      year={2023},
      eprint={2310.06825},
      archivePrefix={arXiv},
      primaryClass={cs.CL}
}

@misc{qwen25technicalreport,
      title={Qwen2.5 Technical Report}, 
      author={Qwen and : and An Yang and Baosong Yang and Beichen Zhang and Binyuan Hui and Bo Zheng and Bowen Yu and Chengyuan Li and Dayiheng Liu and Fei Huang and Haoran Wei and Huan Lin and Jian Yang and Jianhong Tu and Jianwei Zhang and Jianxin Yang and Jiaxi Yang and Jingren Zhou and Junyang Lin and Kai Dang and Keming Lu and Keqin Bao and Kexin Yang and Le Yu and Mei Li and Mingfeng Xue and Pei Zhang and Qin Zhu and Rui Men and Runji Lin and Tianhao Li and Tianyi Tang and Tingyu Xia and Xingzhang Ren and Xuancheng Ren and Yang Fan and Yang Su and Yichang Zhang and Yu Wan and Yuqiong Liu and Zeyu Cui and Zhenru Zhang and Zihan Qiu},
      year={2025},
      eprint={2412.15115},
      archivePrefix={arXiv},
      primaryClass={cs.CL}
}

@article{spearman1987proof,
  title={The proof and measurement of association between two things},
  author={Spearman, Charles},
  journal={The American journal of psychology},
  volume={100},
  number={3/4},
  pages={441--471},
  year={1987},
  publisher={JSTOR}
}

@inproceedings{Zheng2023judge,
author = {Zheng, Lianmin and Chiang, Wei-Lin and Sheng, Ying and Zhuang, Siyuan and Wu, Zhanghao and Zhuang, Yonghao and Lin, Zi and Li, Zhuohan and Li, Dacheng and Xing, Eric P. and Zhang, Hao and Gonzalez, Joseph E. and Stoica, Ion},
title = {Judging LLM-as-a-judge with MT-bench and Chatbot Arena},
year = {2023},
publisher = {Curran Associates Inc.},
address = {Red Hook, NY, USA},
booktitle = {Proceedings of the NeurIPS},
articleno = {2020},
numpages = {29},
location = {New Orleans, LA, USA},
series = {NIPS '23}
}

@article{DBLP:journals/corr/abs-2608-29856,
  author       = {Yifan Chen and
                  Haitao Li and
                  Qingyao Ai and
                  Fengbin Zhu and
                  Tat{-}Seng Chua and
                  Min Zhang and
                  Yiqun Liu},
  title        = {GenRubric: Self-Evolving Rubric Generation for Scalable {LLM} Evaluation},
  journal      = {CoRR},
  volume       = {abs/2608.29856},
  year         = {2026},
  url          = {https://doi.org/10.48550/arXiv.2608.29856},
  doi          = {10.48550/ARXIV.2608.29856}
}
\bibliographystyle{references}

\appendix
\section{Mechanistic Analysis: Experimental Details}

\subsection{dataset}\label{app:data}
\begin{figure}[ht]
    \centering
    \includegraphics[width=0.87\linewidth]{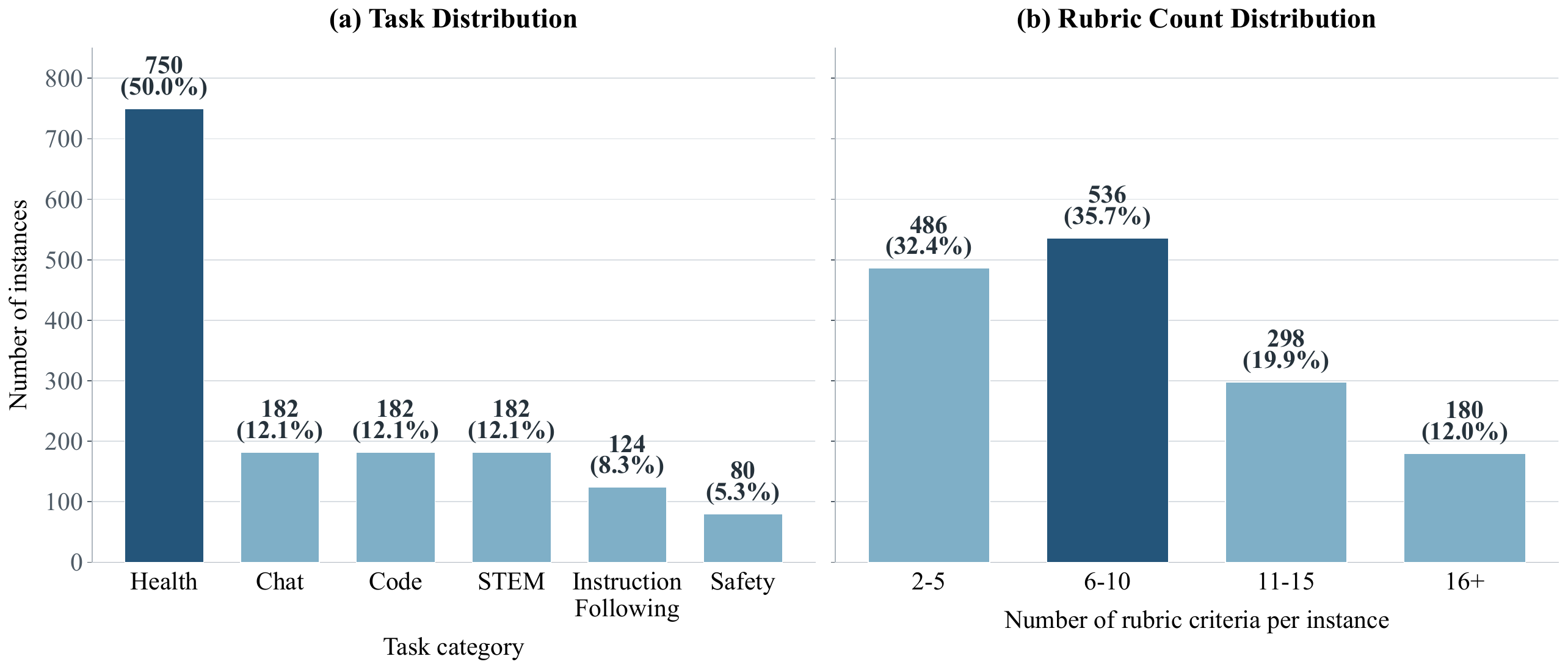}
    \caption{Distribution of task categories and rubric counts in the mechanism analysis set.}
    \label{fig:analysis_distribution}
\end{figure}

The HealthBench dataset~\citep{arora2025healthbench} provides physician-authored criteria for evaluating open-ended medical responses, while RubricBench~\citep{zhou2026rubricbench} covers a broader range of domains, including safety, instruction following, dialogue, STEM, and coding, with expert-annotated atomic criteria, candidate response pairs, and human preference labels. We sample 750 instances from fixed versions of each dataset using stratified sampling. For HealthBench, we perform proportional sampling based on theme combinations and rubric cardinality, preserving the original distributions of task themes and rubric complexity. For RubricBench, we retain all Safety and Instruction Following instances and sample the remaining instances from the Chat, STEM, and Code categories according to their original task domains, data sources, and rubric-count intervals. The resulting mechanistic analysis set comprises 1,500 instances across six task categories. We further summarize the distributions of task categories and rubric counts in Figure~\ref{fig:analysis_distribution}.

\subsection{metric} \label{app:metrics}

\paragraph{Coverage.} For the $i$-th instance, let $G_i$ denote the set of expert-annotated
gold leaves and $R_i$ the set of generated rubric criteria. We define
the coverage indicator as
\begin{equation}
c_{ig} =
\begin{cases}
1, & \text{if the generated rubric fully covers gold leaf } g, \\
0, & \text{otherwise}.
\end{cases}
\end{equation}

The coverage of instance $i$ is computed as
\begin{equation}
\operatorname{Coverage}_i
=
\frac{1}{|G_i|}
\sum_{g \in G_i} c_{ig}.
\end{equation}

\paragraph{Novelty.}
Similarly, the novelty indicator is defined as
\begin{equation}
v_{ir} =
\begin{cases}
1, & \text{if } r \text{ is valid, query-relevant, and absent from } G_i, \\
0, & \text{otherwise}.
\end{cases}
\end{equation}

The valid novelty of instance $i$ is computed as
\begin{equation}
\operatorname{Novelty}_i
=
\frac{1}{|R_i|}
\sum_{r \in R_i} v_{ir}.
\end{equation}

\subsection{Counterfactual response} \label{app:counterfactual}
In this section, we detail the construction of the counterfactual responses. We follow a minimal-intervention principle, modifying only the content required to violate the targeted gold leaves while leaving unrelated content unchanged. The edits include removing required information, altering factual content, violating formatting constraints, omitting necessary safety warnings, modifying applicability conditions, and introducing behaviors specified by negative criteria. After each edit, we re-evaluate the response against all gold leaves and retain it only if the targeted leaves change state while all non-targeted leaves remain unchanged. If a sample fails this isolation check, it is further revised and discarded if the required isolation cannot be achieved. Across the 1,500 instances, this procedure yields 1,498 single-leaf, 1,482 within-dimension multi-leaf, and 1,409 cross-dimension multi-leaf counterfactual responses.

\subsection{Shuffled Path} \label{app:shuffled_path}
For each counterfactual type, we randomly sample 50\% of the instances
for which such mismatched paths can be constructed. We repeat this
procedure with two random seeds, denoted as Seed 1 and Seed 2, which
correspond to 20260813 and 123456, respectively. For a fair comparison, \textsc{Correct Path} and \textsc{Shuffled Path} are evaluated on the same random subset. Consequently, the \textsc{Correct Path} scores vary across seeds and differ from the
full-set results in Table~\ref{tab:counterfactual_tree_comparison}.
\begin{table}[t]
\caption{
Comparison between correct and shuffled hierarchical paths. Values in parentheses denote changes from \textsc{Correct} to \textsc{Shuffled}.
}
\label{tab:shuffled_path_comparison}
\centering
\setlength{\tabcolsep}{5pt}
\begin{tabular}{lcccc}
\toprule
& \multicolumn{2}{c}{Seed 1}
& \multicolumn{2}{c}{Seed 2} \\
\cmidrule(lr){2-3}
\cmidrule(lr){4-5}
Counterfactual Type
& Correct & Shuffled
& Correct & Shuffled \\
\midrule

\multicolumn{5}{l}{\textbf{DeepSeek-V4-Flash-GA}} \\

Single-leaf
& \textbf{23.77}
& 23.63 \loss{-0.13}
& 21.23
& \textbf{21.50} \gain{0.27} \\

Within-dimension
& \textbf{12.55}
& 11.61 \loss{-0.94}
& \textbf{12.28}
& 11.88 \loss{-0.40} \\

Cross-dimension
& \textbf{27.66}
& 25.67 \loss{-1.99}
& \textbf{27.09}
& 25.11 \loss{-1.99} \\

\midrule
\multicolumn{5}{l}{\textbf{Llama3.1-405B-Instruct}} \\

Single-leaf
& \textbf{35.91}
& 34.05 \loss{-1.87}
& \textbf{37.52}
& 36.98 \loss{-0.53} \\

Within-dimension
& \textbf{14.71}
& 13.09 \loss{-1.62}
& \textbf{15.65}
& 13.77 \loss{-1.89} \\

Cross-dimension
& \textbf{43.26}
& 42.41 \loss{-0.85}
& \textbf{44.26}
& 42.55 \loss{-1.70} \\

\bottomrule
\end{tabular}
\end{table}

The experimental results are reported in Table~\ref{tab:shuffled_path_comparison}. The correct path generally outperform their shuffled counterparts across judge models, random seeds, and counterfactual types. This suggests that the gains from path conditioning cannot be attributed to additional context length alone. This indicates that the judge benefits specifically from the structured semantics encoded by the hierarchy. 

\subsection{Calibration-Based Selection for RQ2}
\label{app:rq2_calibration}

For each local candidate $C$, we use two calibration judges to independently score every criterion on the same set of six calibration responses using a five-point scale. Based on these scores, we characterize each candidate set from four aspects, including response discrimination, cross-judge variance, empirical redundancy, and the number of criteria. Response discrimination captures how much the criterion scores vary across calibration responses and is measured by their average variance. Cross-judge variance reflects the extent to which the two judges disagree on the same criterion-response pair. Finally, the empirical redundancy captures similarity in scoring behavior across criteria and is defined as the proportion of criterion pairs whose response-score profiles have a Pearson correlation above 0.90. We combine these signals into a calibration utility that rewards response discrimination while penalizing cross-judge variance, empirical redundancy, and unnecessary growth in the number of criteria, as defined below.
\begin{equation}
U_{\mathrm{RQ2}}(C)
=
D_{\mathrm{RQ2}}(C)
-
V_{\mathrm{RQ2}}(C)
-
0.10R_{\mathrm{RQ2}}(C)
-
0.02\max(0,|C|-1).
\label{eq:rq2_calibration_utility}
\end{equation}
For \textsc{Semantic+Calibration}, this utility is applied only after semantic filtering. A candidate replaces the default semantic choice only when its utility improves by at least $0.05$. The calibration responses are used exclusively for candidate selection, while the discrimination, judge variance, and pairwise accuracy reported in Table~\ref{tab:rq2_semantic_calibration} are computed independently on the held-out response pairs.

\section{ARTS-Hybrid: Experimental Details}

\subsection{calibration objective}\label{app:calibration_objective}
For a semantically feasible decomposition $C$ of the current node, let $c\in C$ denote a child criterion. Each criterion is independently evaluated by $n=2$ calibration judges on the same set of $m=6$ calibration responses. We define
\begin{equation}
p_{c,r,j}\in[0,1],
\label{eq:calibration_score}
\end{equation}
where $p_{c,r,j}$ denotes the probability assigned by judge $j$ that response $r$ satisfies criterion $c$.

\paragraph{Candidate-level score aggregation.}
To characterize the behavior of a candidate decomposition $C$ on each response, we first aggregate scores across its child criteria using a soft-min operator as
\begin{equation}
z_{r,j}(C)
=
\left(
\frac{1}{|C|}
\sum_{c\in C}
\max(p_{c,r,j},\epsilon)^{-\beta}
\right)^{-1/\beta},
\label{eq:softmin_aggregation}
\end{equation}
where $\epsilon=10^{-6}$ and $\beta=8$. This aggregation approximates the minimum score across child criteria, such that low satisfaction on any criterion reduces the overall score of the candidate decomposition. When $C$ contains a single criterion, we directly set $z_{r,j}(C)=p_{c,r,j}$. 

Based on the aggregated scores $z_{r,j}(C)$, we next define the response discrimination $D(C)$, judge agreement $A(C)$, score stability $S(C)$, and behavioral redundancy $R(C)$ that constitute the calibration objective in Equation~\ref{eq:calibration_objective}.

\paragraph{Response discrimination.}
The response discrimination measures whether a candidate decomposition produces sufficiently varied scores across the calibration responses. We compute the variance of the aggregated scores across responses for each judge and then average over judges as
\begin{equation}
D(C)
=
\frac{1}{n}
\sum_{j=1}^{n}
\operatorname{Var}_{r}
\left[
z_{r,j}(C)
\right].
\label{eq:response_discrimination}
\end{equation}
The larger $D(C)$ indicates greater variation across responses and thus stronger discriminative power.

\paragraph{Cross-judge agreement.}
The cross-judge agreement measures whether different calibration judges assign similar scores to the same criterion-response pair. For each criterion $c$, we first compute the score variance across judges for each response as
\begin{equation}
v_{c,r}
=
\operatorname{Var}_{j}
\left[
p_{c,r,j}
\right].
\label{eq:cross_judge_variance}
\end{equation}
We then define the agreement of criterion $c$ as
\begin{equation}
a(c)
=
\frac{1}{
1+\frac{1}{m}\sum_{r=1}^{m}v_{c,r}
}.
\label{eq:criterion_agreement}
\end{equation}
The overall agreement of a candidate decomposition is obtained by averaging across its child criteria:
\begin{equation}
A(C)
=
\frac{1}{|C|}
\sum_{c\in C}
a(c).
\label{eq:cross_judge_agreement}
\end{equation}
The larger $A(C)$ indicates greater agreement between judges on the same responses and criteria.

\paragraph{Judge stability.}
Beyond agreement on individual responses, the judge stability measures whether different judges exhibit similar discrimination behavior across the calibration responses. For each judge $j$, we first compute the variance of the aggregated scores across responses as
\begin{equation}
u_j(C)
=
\operatorname{Var}_{r}
\left[
z_{r,j}(C)
\right].
\label{eq:judge_profile_variance}
\end{equation}
We then define judge stability as
\begin{equation}
S(C)
=
\frac{1}{
1+\operatorname{Var}_{j}
\left[
u_j(C)
\right]
}.
\label{eq:judge_stability}
\end{equation}
While $A(C)$ captures agreement on individual responses, $S(C)$ captures the consistency of discrimination patterns across judges, providing a complementary measure of overall scoring stability.

\paragraph{Score redundancy.}
The score redundancy measures the similarity of scoring patterns across child criteria on the calibration responses. We first average each criterion's scores across judges as
\begin{equation}
\bar{p}_{c,r}
=
\frac{1}{n}
\sum_{j=1}^{n}
p_{c,r,j}.
\label{eq:mean_criterion_score}
\end{equation}
For each criterion $c$, let
\begin{equation}
\bar{\mathbf{p}}_c
=
\left[
\bar{p}_{c,1},
\ldots,
\bar{p}_{c,m}
\right]
\label{eq:criterion_score_profile}
\end{equation}
denote the score vector across calibration responses. We use Spearman rank correlation~\citep{spearman1987proof} to measure the similarity between the scoring profiles of each pair of child criteria, and define \(R(C)\) as the maximum positive correlation
\begin{equation}
R(C)
=
\max_{\substack{c,c'\in C}}
\max
\left(
0,\,
\rho_{\mathrm{S}}
\left(
\bar{\mathbf{p}}_c,
\bar{\mathbf{p}}_{c'}
\right)
\right).
\label{eq:score_redundancy}
\end{equation}
When $|C|=1$, we set $R(C)=0$. The larger $R(C)$ indicates stronger similarity in scoring behavior among child criteria, suggesting increased redundancy.

\subsection{Implementation Details}
\label{app:implementation}

We use GLM-4.7-Flash as the common rubric generator and judge model across all evaluation methods. For ARTS-Hybrid, we follow the response construction and calibration protocol introduced in RQ2, using the same calibration panel and judge models. The specific models used for calibration response generation and judging are listed in Table~\ref{tab:calibration_models}.

\begin{table}[ht]
\caption{Calibration response generators and judges used in
ARTS-Hybrid. Each response generator produces two responses, and each
calibration judge evaluates all six responses.}
\label{tab:calibration_models}
\centering
\setlength{\tabcolsep}{4pt}
\begin{tabular}{lll}
\toprule
Role & Target Quality & Model \\
\midrule
Response Generator
& Strong
& Qwen3-30B-A3B-Instruct-2507 \\

Response Generator
& Medium
& Llama-3.1-70B-Instruct \\

Response Generator
& Weak
& Mistral-7B-Instruct-v0.2 \\
\midrule
Calibration Judge
& --
& Qwen2.5-72B-Instruct \\

Calibration Judge
& --
& Llama-3.3-70B-Instruct \\
\bottomrule
\end{tabular}
\end{table}

\subsubsection{Response-Based Calibration Setup} \label{app:calibration}
\begin{figure}[ht]
    \centering
    \includegraphics[width=0.9\linewidth]{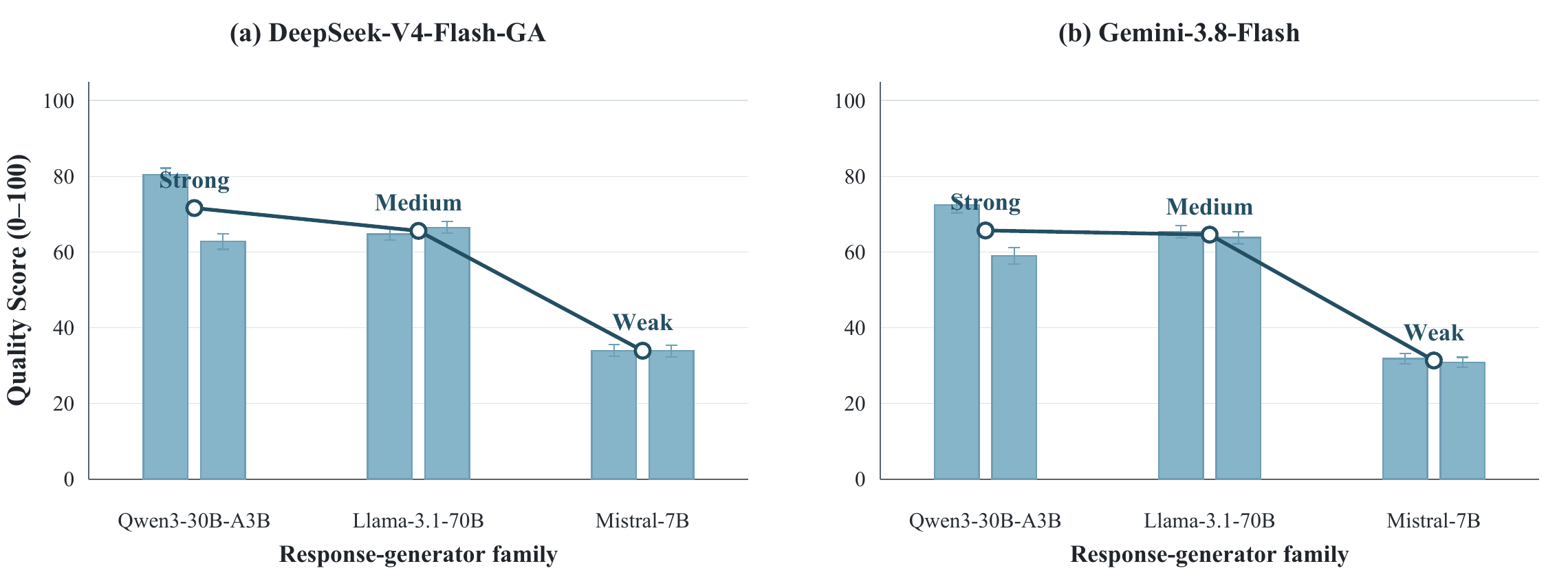}
    \caption{Independent validation of calibration response quality. DeepSeek-V4-Flash-GA and Gemini-3.8-Flash independently evaluate responses generated at the intended strong, medium, and weak quality levels.}
    \label{fig:calibration_quality}
\end{figure}

\paragraph{Calibration responses.}
For each query, we pre-generate six calibration responses before rubric construction and keep them fixed throughout the tree search. \textbf{The same calibration panel is shared across all candidate criteria.} To cover diverse response quality and generation characteristics, we use three model families, with two responses generated by each model. Specifically, we use Qwen3-30B-A3B-Instruct-2507~\citep{yang2025qwen3} to generate two strong responses, Llama-3.1-70B-Instruct to generate two medium-quality responses, and Mistral-7B-Instruct-v0.2~\citep{jiang2023mistral7b} to generate two weak responses. We further control the intended quality level through group-specific generation instructions. For the strong group, the instructions emphasize accuracy, completeness and adherence to explicit requirements, with a variant encouraging greater conciseness. For the medium group, the instructions require the response to address the main task while allowing minor omissions or incomplete explanations, with a variant encouraging greater elaboration. For the weak group, the instructions preserve overall plausibility while allowing incomplete, vague, or poorly organized content, with a variant introducing a minor factual error. Therefore, these responses provide a common calibration basis spanning different quality levels and response characteristics.

To verify that the calibration responses follow the intended quality levels, we use two strong independent evaluators, DeepSeek-V4-Flash-GA and Gemini-3.8-Flash, to assess their response quality. As shown in Figure~\ref{fig:calibration_quality}, both evaluators exhibit a consistent quality gradient, with average scores decreasing from the strong to medium and weak response groups. This pattern aligns with the intended quality ordering and supports the use of these responses as a quality-diverse calibration panel.

\paragraph{Calibration judges.} We use Qwen2.5-72B-Instruct and Llama-3.3-70B-Instruct as two independent calibration judges. Each judge evaluates the candidate criteria on the same calibration responses, providing the scoring signals used for response-based candidate selection. The calibration responses are used exclusively during rubric construction and are excluded from the final evaluation.

% All experiments are conducted on 16 NVIDIA H200 GPUs across two nodes.

\begin{table}[ht]
\caption{The hyperparameter configurations used by ARTS-Hybrid.}
\label{tab:arts_hyperparameters}
\centering
\setlength{\tabcolsep}{6pt}
\begin{tabular}{lc}
\toprule
Hyperparameter & Value \\
\midrule
Leaf budget & 16 \\
Candidates per expansion & \{6, 10\} \\
Minimum children per expansion & \{2, 3\} \\
Maximum tree depth & 3 \\
Depth discount $\gamma$ & 0.85 \\
\midrule
Calibration responses per query & 6 \\
Calibration judges & 2 \\
Semantic score threshold & 0.60 \\
Calibration threshold $\tau_{\mathrm{scor}}$ & \{0.75, 0.80\} \\
Coverage threshold $\tau_{\mathrm{cov}}$ & \{0.80, 1.00\} \\
Maximum regeneration attempts & \{2, 4\} \\
\midrule
Objective weights $(\alpha_D,\alpha_A,\alpha_S)$
& (0.25, 0.30, 0.45) \\
Redundancy penalty $\lambda_{\mathrm{red}}$ & \{0.05, 0.10\} \\
Size penalty $\lambda_{\mathrm{size}}$ & \{0.00, 0.01\} \\
Maximum global repair rounds & 2 \\
\bottomrule
\end{tabular}
\end{table}

\subsubsection{Hyperparameters} \label{app:params}
In this section, we summarize the hyperparameter configurations used in ARTS-Hybrid, including the search settings for rubric tree construction, the thresholds and regeneration settings used during semantic and calibration-based selection, and the weights and regularization terms in the calibration objective. Table~\ref{tab:arts_hyperparameters} reports the corresponding values used in our experiments.

\subsubsection{Rubric Generator Distillation}
We fine-tune Qwen3-4B on the same stratified set of 3,000 OpenRubrics instances using rubrics from different supervision sources. All variants use identical training examples and hyperparameters. We apply LoRA with rank 16, a scaling factor of 32, and dropout of 0.05 for one epoch, with a learning rate of $2\times10^{-5}$, a global batch size of 32, and a maximum sequence length of 4,096. Training is conducted on 8 NVIDIA H200 GPUs. For evaluation on JudgeBench, we use GLM-4.7-Flash as the common response judge and to resolve ties in rubric-aggregated scores.

\subsubsection{Rubric-as-Reward Training Details}
We initialize all policy optimization experiments from the same Qwen3-4B model and use a fixed stratified set of 3,000 OpenRubrics prompts. For each rollout, a frozen Qwen2.5-7B-Instruct model independently evaluates whether the response satisfies each rubric criterion. The resulting binary criterion-level judgments are aggregated using normalized criterion weights to produce a scalar reward. We optimize the policy with Dr.GRPO for three epochs, corresponding to 186 optimization steps. Each step contains 48 prompts with eight rollouts per prompt. We use an actor learning rate of $1\times10^{-6}$, maximum prompt and response lengths of 4,096 tokens, and a low-variance KL penalty with a coefficient of 0.001. Following Dr.GRPO, we disable group-reward standard-deviation normalization and use sequence-level token-sum loss aggregation. All reward-source variants use identical policy initialization, training prompts, graders, and optimization settings. The thinking and non-thinking settings differ only in whether thinking mode is enabled. For evaluation on HealthBench~\citep{arora2025healthbench}, we use Qwen2.5-7B-Instruct as the judge model to evaluate each generated response against the physician-authored criteria and aggregate the criterion-level judgments into the final HealthBench score.

\section{Additional Results} \label{app:additional_results}

\paragraph{Cross-Evaluator rubric quality.}
\label{app:intrinsic_quality}

To assess robustness across evaluators, we repeat the intrinsic rubric quality evaluation using Llama3.1-405B-Instruct as an additional independent evaluator. As shown in Table~\ref{tab:intrinsic-rubric-quality-1}, the ARTS-Hybrid achieves the highest overall score and performs best in concreteness, non-redundancy, and actionability.

\begin{table}[ht]
\caption{
Comparison of intrinsic rubric quality. Higher values indicate better performance.
}
\label{tab:intrinsic-rubric-quality-1}
\centering
\setlength{\tabcolsep}{4pt}
\begin{tabular}{lccccc}
\toprule
Method
& Coverage $\uparrow$
& Concreteness $\uparrow$
& Non-red. $\uparrow$
& Actionability $\uparrow$
& Overall $\uparrow$ \\
\midrule

Query-Only
& 3.679
& 4.251
& 3.420
& 4.219
& 3.609 \\

Response-Aware
& 3.711
& 4.262
& 3.248
& 4.216
& 3.600 \\

Q-World
& \textbf{4.478}
& 4.047
& 1.948
& 4.064
& 3.119 \\

ARTS-Hybrid (ours)
& 4.009
& \textbf{4.405}
& \textbf{4.073}
& \textbf{4.402}
& \textbf{3.938} \\

\bottomrule
\end{tabular}
\end{table}

\paragraph{Training dynamics of rubric-based rewards.}

\begin{figure}[ht]
    \centering
    \includegraphics[width=\linewidth]{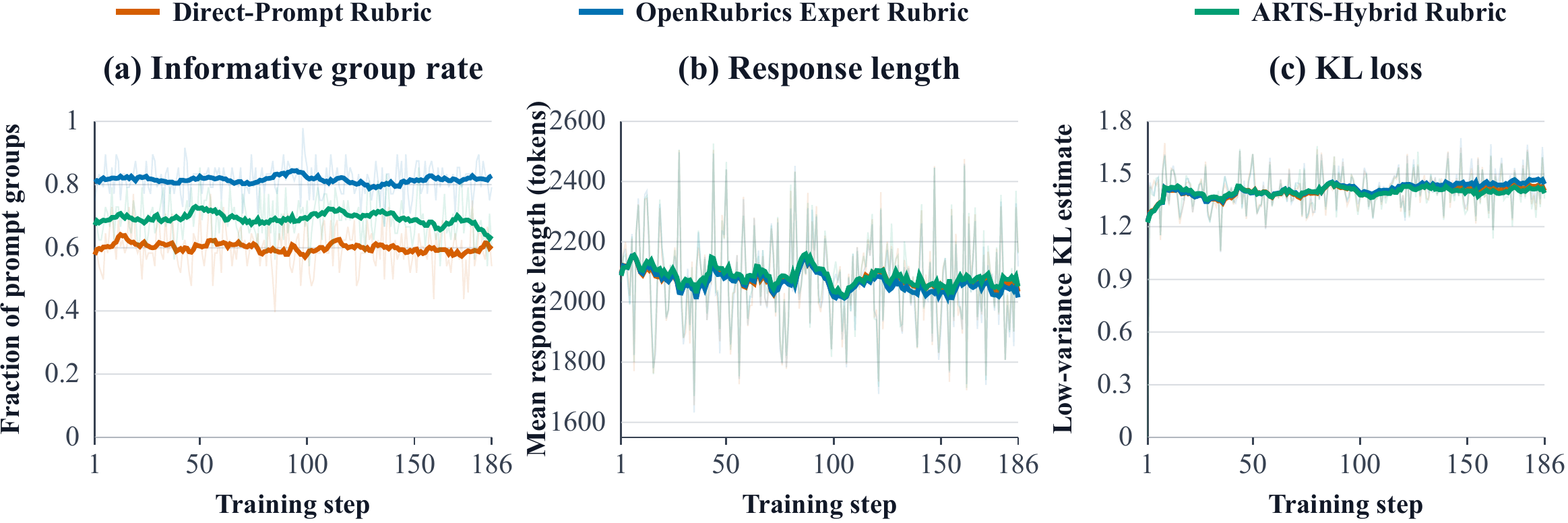}
    \caption{Training dynamics under different rubric-based reward signals, including informative group rate, response length, and KL loss.}
    \label{fig:rl_training_dynamics}
\end{figure}

To further examine how different rubric-based rewards affect policy optimization, we track three diagnostic metrics throughout Dr.GRPO training. Figure~\ref{fig:rl_training_dynamics}(a) reports the \textit{Informative Group Rate}, defined as the fraction of prompt groups whose rollout rewards are not identical and therefore yield non-zero GRPO advantages. ARTS-Hybrid provides informative relative reward signals for more prompt groups than Direct-Prompt, while OpenRubrics Expert achieves the highest rate. Notably, ARTS-Hybrid still yields better downstream performance than OpenRubrics Expert, suggesting that reward effectiveness depends not only on how often rewards differentiate responses, but also on \textbf{whether these differences reflect task-relevant response quality}. Figure~\ref{fig:rl_training_dynamics}(b) shows the mean response length during training. The three curves closely overlap and remain stable at approximately 2,000-2,100 tokens, indicating that the gains from ARTS-Hybrid are not driven by a preference for longer responses. Finally, Figure~\ref{fig:rl_training_dynamics}(c) shows similar and stable KL loss across all three methods, suggesting that the advantage of ARTS-Hybrid does not arise from more aggressive policy updates.

\paragraph{Cross-judge robustness.} 
In addition, we examine whether the gains of ARTS-Hybrid are sensitive to the choice of judge model. As shown in Table~\ref{tab:judge-model-comparison}, the ARTS-Hybrid achieves the highest pairwise accuracy with both Qwen3.5-397B-A17B and Llama3.1-405B-Instruct~\citep{grattafiori2024llama}, with particularly clear gains under the latter. Notably, while the performance of one-shot rubric generation varies considerably across judges, the ARTS-Hybrid maintains consistent improvements over direct judgment. These results suggest that the effectiveness of the generated rubrics is not tied to a specific evaluator and can generalize across different judge models.
\begin{table}[ht]
\caption{
Pairwise accuracy (\%) across judge models.
The best result is highlighted in bold.
}
\label{tab:judge-model-comparison}
\centering
\setlength{\tabcolsep}{5pt}
\begin{tabular}{lcc}
\toprule
Method
& Qwen3.5-397B-A17B
& Llama3.1-405B-Instruct \\
\midrule
Direct Judge
& 77.14
& 57.40 \\

Query-Only
& 80.57
& 41.50 \\

Response-Aware
& 79.43
& 56.00 \\

Q-World
& 80.00
& 63.86 \\

ARTS-Hybrid (ours)
& \textbf{85.67}
& \textbf{69.71} \\
\bottomrule
\end{tabular}
\end{table}

\paragraph{Ablation study.}
We conduct an ablation study to examine the contributions of semantic validation and response-based calibration. The results are reported in Table~\ref{tab:arts-ablation}. When semantic validation is removed, the performance decreases across all four benchmarks, with the largest drop on the RM-Bench Chat dataset. When response-based calibration is removed, performance decreases on RM-Bench Chat, PreciseIF, and JudgeBench datasets, while remaining comparable on Focus. Overall, both components contribute to the performance of ARTS-Hybrid, with semantic validation showing a more consistent effect across benchmarks.

\begin{table}[ht]
\caption{
Ablation study of semantic specification and response-based calibration.
All results are pairwise evaluation accuracies (\%). The best result
in each column is highlighted in bold.
}
\label{tab:arts-ablation}
\centering
\setlength{\tabcolsep}{5pt}
\begin{tabular}{lcccc}
\toprule
Method
& RM-Bench Chat
& PreciseIF
& Focus
& JudgeBench \\
\midrule
ARTS-Hybrid
& \textbf{64.34}
& \textbf{55.63}
& 85.59
& \textbf{64.29} \\

\quad w/o Semantic
& 50.39
& 54.17
& 77.04
& 61.71 \\

\quad w/o Calibration
& 58.14
& 55.00
& \textbf{86.13}
& 62.86 \\
\bottomrule
\end{tabular}
\end{table}

\section{Prompts}
\label{app:prompts}

In this section, we provide the core prompts used in the mechanistic analysis and ARTS-Hybrid. We omit auxiliary prompts for formatting, retries, and other implementation-specific operations.

\subsection{Prompts for Mechanistic Analysis}

We first provide the prompts used in the mechanistic analysis for RQ1. These include the flat and hierarchical rubric generation prompts used in Table~\ref{tab:tree_flat_comparison}, as well as the criterion judgment prompts used to evaluate counterfactual responses in Table~\ref{tab:counterfactual_tree_comparison}.

\paragraph{Rubric generation.}
For flat rubric generation, we use the following prompt.

\begin{promptbox}{Flat Rubric Generation}
Using ONLY the query, directly create a flat set of atomic evaluation rubrics.

QUERY
<QUERY>

Each item must be independently judgeable, relevant, non-overlapping, and describe what a good response must satisfy. Do not create a hierarchy, inspect or speculate about candidate responses, or reveal an answer.

Return the result in the following format.

{"rubrics": [
  {"criterion": "...", "category": "..."}
]}

Generate 3 to <MAX_LEAVES> items, stopping when the query is adequately covered.
\end{promptbox}

For hierarchical rubric generation proceeds in two stages. We first construct the initial branches of the rubric tree and then recursively examine each node to determine whether further decomposition is needed. The prompts for these two stages are provided below.
\begin{promptbox}{Hierarchical Rubric Generation}
Create the first level of an adaptive evaluation-rubric tree using ONLY the query below.

QUERY
<QUERY>

The root is "Overall response quality" Split it into 2-8 distinct and useful semantic branches. Each branch should describe what a good response must satisfy without revealing the answer. Do not inspect or speculate about candidate responses.

Return the result in the following format.

{"children": [
  {"name": "...", "criterion": "..."}
]}
\end{promptbox}

\begin{promptbox}{Recursive Node Expansion}
Decide whether the current rubric node should remain atomic or be further decomposed, using only the query and its semantic path.

QUERY
<QUERY>

NODE PATH
<NODE_PATH>

CURRENT CRITERION
<CRITERION>

CURRENT DEPTH
<DEPTH>

MAXIMUM DEPTH
<MAX_DEPTH>

Keep the node atomic when it represents one independently judgeable requirement or failure mode. Split it only when it contains multiple non-overlapping requirements that would be more diagnostic if assessed separately. Avoid unary splits and cosmetic restatements.

Return either an atomic decision or a split containing 2-6 child
criteria.
\end{promptbox}

\paragraph{Counterfactual violation detection.}
% Under the \textsc{Flat} condition, each leaf is judged independently.

% \begin{promptbox}{Flat Criterion Judgment}
% Judge ONE Gold rubric leaf without hierarchical context. Make the final pass/fail decision only for this leaf criterion.

% For desired_state=criterion_met, pass means that the response satisfies the criterion. For desired_state=criterion_not_met, pass means that the undesirable behavior is absent. Use only observable response content.

% QUERY
% <QUERY>

% RESPONSE
% <RESPONSE>

% LEAF CRITERION
% <CRITERION>

% DESIRED STATE
% <DESIRED_STATE>

% Return the result in the following format.

% {"passes": true, "evidence": "..."}
% \end{promptbox}

We provide the core prompt used for counterfactual violation detection under the \textsc{Path} condition. The root-to-leaf path provides semantic context for judging the final leaf criterion, while intermediate nodes are not used as routing gates.

\begin{promptbox}{Path-Conditioned Criterion Judgment}
You are given ONE root-to-leaf semantic path from a larger Gold rubric tree. The final item in the path is the only leaf being judged.

Use the parent nodes to understand the leaf's intended semantic scope, terminology, and required evidence. Parent nodes are context rather than separately scored routing gates.

QUERY
<QUERY>

RESPONSE
<RESPONSE>

FULL SEMANTIC PATH
<SEMANTIC_PATH>

LEAF CRITERION
<CRITERION>

DESIRED STATE
<DESIRED_STATE>

Determine whether the response satisfies the final leaf criterion.

Return the result in the following format.

{"passes": true, "evidence": "..."}
\end{promptbox}

We next provide the core prompts used in RQ2, where we examine how semantic assessment and response-based calibration provide complementary signals for criterion selection.

\paragraph{Semantic and calibration-based criterion selection.}

The Semantic Assessment Prompt determines whether a local revision is semantically better than, or at least no worse than, the original Direct Rubric. The Calibration Scoring Prompt evaluates how candidate criteria behave on actual responses. As described above, calibration-based candidate selection is performed programmatically using the utility defined in Equation~\ref{eq:rq2_calibration_utility}, rather than through an additional model call. We therefore provide only the Calibration Scoring Prompt below.

\begin{promptbox}{Semantic Assessment}
You are an independent semantic rubric comparator. Compare an immutable Direct rubric set with one local revision candidate.

QUERY
<QUERY>

DIRECT RUBRIC
<DIRECT_RUBRIC>

CANDIDATE REVISION
<CANDIDATE>

Compare the two alternatives along four dimensions.

1. Coverage of material requirements grounded in the query.
2. Atomicity of independently fail-able requirements.
3. Concrete and observable scoreability.
4. Absence of true semantic duplication or subsumption.

For each dimension, decide whether Direct is better, Candidate is better, or they are tied. Identify any requirement that is lost, weakened, broadened, or duplicated by the candidate.

Do not use Gold rubrics, responses, calibration scores, or preference labels.

Return the comparison for all four dimensions, any candidate regressions, and a brief summary.
\end{promptbox}

\begin{promptbox}{Calibration Scoring}
Score every calibration response against the ONE rubric below independently.

QUERY
<QUERY>

RUBRIC
<RUBRIC>

CALIBRATION RESPONSES
<CALIBRATION_RESPONSES>

For desired_state=criterion_met, a high score means that the response satisfies the criterion. For desired_state=criterion_not_met, a high score means that the undesirable behavior is absent.

Use the following scale.

1 means clearly violating the desired state.
2 means mostly violating it.
3 means mixed, partial, or genuinely borderline.
4 means mostly satisfying it with a minor defect.
5 means clearly and fully satisfying it.

Judge only observable response content. Do not use the response's declared quality target as a substitute for inspection.

Return one score for each calibration response.
\end{promptbox}

We do not provide a separate prompt for RQ3, as the same judging protocol is used across all settings, with criterion-level judgments held fixed while only the score aggregation scheme is varied.

\subsection{Prompts for ARTS-Hybrid}

In this section, we provide the core prompts used in ARTS-Hybrid. We first present the node decomposition prompt used to generate candidate child criteria, followed by the semantic validation prompt used to identify semantically feasible candidates. Calibration-based selection does not require an additional prompt, as feasible candidates are selected programmatically using the calibration objective defined above. Once the tree search is complete, ARTS-Hybrid performs global validation over the resulting leaf set, for which we also provide the corresponding prompt below.

% \paragraph{Candidate Decomposition.}
% ARTS-Hybrid generates candidate child criteria conditioned on the current node. For a non-root node, we use the following prompt.

\begin{promptbox}{Candidate Decomposition}
USER QUERY
<QUERY>

PARENT RUBRIC
<PARENT_RUBRIC>

Generate <N_CANDIDATES> sub-rubrics that jointly preserve and cover the meaning of the parent rubric for this query.

Each sub-rubric must be concrete, verifiable, independently meaningful, specific to the query, and sufficiently unambiguous for different judges to interpret consistently. Avoid redundant near-duplicates.

Use AND semantics. Every selected child should represent a necessary aspect of the parent. Do not assign weights.

Return the selected sub-rubrics.
\end{promptbox}

% \paragraph{Semantic Validation.}
% Each generated criterion is first evaluated for semantic validity.

\begin{promptbox}{Semantic Validation}
Evaluate whether a rubric criterion is suitable for this exact task.

USER QUERY
<QUERY>

CRITERION
<CRITERION>

Assess the following properties on a 0-1 scale.

- concrete means specific, observable, and not vague.
- verifiable means answerable as YES/NO from the response.
- unambiguous means likely to be interpreted consistently across judges.
- task_specific means tied to the query rather than generic quality.
- distinctive means capable of distinguishing responses of different quality.
- leaf_quality means suitable as a final scoring criterion.
- atomic means containing exactly one independently fail-able requirement.

Also determine whether the criterion is substantive, format-only, and an explicit query requirement.

A criterion is semantically admissible only if it is concrete, verifiable, unambiguous, and task-specific. A valid but non-atomic criterion may be retained as an internal node but not as a final leaf.

Return the assessment for all properties together with the final decision and a brief reason.
\end{promptbox}

% Semantically admissible criteria are subsequently assembled into feasible decompositions. Calibration-based ranking among feasible decompositions is performed programmatically using Equation~\ref{eq:calibration_objective}.

% \paragraph{Global Validation.}
% Once tree construction is complete, we validate the final leaf set using the following prompt.

\begin{promptbox}{Global Validation}
Check whether the final leaf rubric set adequately covers the evaluation needs of this task.

USER QUERY
<QUERY>

ROOT RUBRIC
<ROOT_RUBRIC>

FINAL LEAF RUBRICS
<LEAF_RUBRICS>

The leaf set must satisfy the following requirements.

- Cover the principal task-specific requirements.
- Preserve every explicit query constraint.
- Express independently fail-able requirements as separate leaves.
- Contain concrete and directly scoreable criteria.
- Avoid semantic duplication and strict subsumption.
- Include substantive correctness, reasoning, or domain criteria when the task contains substantive content.

If validation fails, identify the missing aspects, atomicity violations, and redundant pairs.

Return the validation decision together with the corresponding diagnostics.
\end{promptbox}

\section{Case Study of ARTS-Hybrid Rubric Construction}
\label{app:case_study}

We present a representative example to illustrate how ARTS-Hybrid structures and refines evaluation criteria compared with direct rubric generation.

\begin{casebox}{Query}

What is the author's purpose for writing this article? What words or phrases indicate that purpose? How does this text connect back to Chapter One and the Prologue of \textit{Hidden Figures}?

\end{casebox}

\begin{casebox}{Direct-Generated Rubric}

\textbf{D1.} Identify the author's primary purpose.

\textbf{D2.} Provide at least two words or phrases indicating that purpose.

\textbf{D3.} Explain how the text connects to both Chapter One and the Prologue.

\textbf{D4.} Write the response in complete sentences.

\end{casebox}

\begin{casebox}{ARTS-Hybrid-Generated Rubric}

\textbf{A1.} Identify the author's purpose for writing the article.

\textbf{A2.} Cite specific words or phrases that indicate this purpose.

\textbf{A3.} Explain how the article connects to Chapter One.

\textbf{A4.} Support the Chapter One connection with details from the article.

\textbf{A5.} Explain how the article connects to the Prologue.

\textbf{A6.} Use relevant evidence to support the cross-text connections.

\end{casebox}

Compared with direct generation, the ARTS-Hybrid produces finer-grained criteria that separate independently assessable requirements. The directly generated rubric combines the connections to Chapter One and the Prologue into a single criterion, while introducing additional constraints such as requiring at least two supporting phrases and complete sentences.

\end{document}